\documentclass[runningheads]{llncs}

\usepackage[T1]{fontenc}
\usepackage{lmodern}
\usepackage{graphicx}
\usepackage{booktabs}
\usepackage{array}
\usepackage{multirow}
\usepackage{amsmath,amssymb}
\usepackage{microtype}
\usepackage{tikz}
\usetikzlibrary{arrows.meta,positioning}
\usepackage[colorlinks=true,linkcolor=blue!50!black,citecolor=blue!50!black,urlcolor=blue!50!black]{hyperref}
\usepackage{placeins}
\usepackage{rotating}
\makeatletter
\renewcommand{\definition}{\@spthm{definition}{\definitionname}{\bfseries}{\upshape}}
\makeatother

\usepackage{xcolor}
\begin{document}

\title{FERRETS: A Scalable Symbolic Transform for Time Series
Classification}
\titlerunning{FERRETS: A Symbolic Transform for Time Series Classification}
\author{Antoine Guillaume}
\authorrunning{A. Guillaume}
\institute{Novahe, Constellation Group, France\\
\email{antoine.guillaume@novahe.fr}}

\maketitle

\begin{abstract}
Dictionary methods for time series build bags of symbolic words from
subsequences, but the accurate ones scale poorly. Against other
families of methods, their accuracy no longer pays for their cost. We present
FERRETS, a label-free dictionary transform built on WEASEL~2. It keys windows
of random length and dilation into exact eight-bit words, adds a level word of
each window's mean and spread, and on multivariate data adds cross-channel
words. We compare it with leading classifiers, deep networks and foundation
models included, on the UCR, Multiverse and MONSTER archives. FERRETS is the
fastest of the eleven methods timed on UCR under a ten-thread budget. Of the ten others, only
MultiRocket-Hydra is significantly more accurate, taking
24.7 times as long. FERRETS is not significantly different from WEASEL~2, the
most accurate earlier dictionary method, which takes 125
times as long, and is significantly ahead of the other dictionary methods. On
the Multiverse core, it is not significantly different in accuracy from HIVE-COTE~2 and
MultiRocket-Hydra, though behind both on UCR. It is also significantly more
accurate than five frozen foundation models on UCR's published split. On the MONSTER problems both
complete, MiniRocket takes 3.7 times as long to fit as
the streaming FERRETS classifier, which is significantly more accurate.

\keywords{Time series transform \and Time series classification \and Symbolic
representations \and Dictionary methods \and Scalability.}
\end{abstract}

\section{Introduction}
\label{sec:intro}

Time series classification has long been benchmarked on the UCR
archive~\cite{UCRArchive}, whose problems are small.
The MONSTER archive~\cite{MONSTER}, with problems of tens of thousands to tens of
millions of series, and the multivariate Multiverse archive~\cite{multiverse}
show a limit of UCR, since many published methods cannot process such data
within a practical time or memory budget (Section~\ref{sec:cost}). An ideal
classifier for such data must fit in time linear in the number and length of
the training series, and predict at a cost per series that does not grow with
the training set. Its fit must be able to stream the training set in passes
rather than hold it in memory, and its cost must stay within budget on series
with hundreds of channels. Finally, it must meet these limits at an accuracy
close to that of the methods that do not.

Dictionary methods classify a series from the counts of its \emph{words}, short
symbol strings that stand for discretized windows (Section~\ref{sec:definitions}).
Unlike a convolution feature, as in MiniRocket~\cite{MiniRocket}, a word
can be decoded and located in a series. However, the accurate dictionary methods
are costly by design (Section~\ref{sec:sota-dict}). The nearest-neighbor ones,
such as BOSS~\cite{BOSS} and TDE~\cite{tde}, compare a test series with every
training series. The linear ones, notably WEASEL~2 as implemented in
aeon~\cite{schafer2023weasel,aeon} and MrSQM~\cite{mrsqm}, keep the words that a
test against the labels retains, and must
first hold the counts of every candidate word on the whole training set in
memory.

\begin{figure}[tb]
\centering
\includegraphics[width=\textwidth]{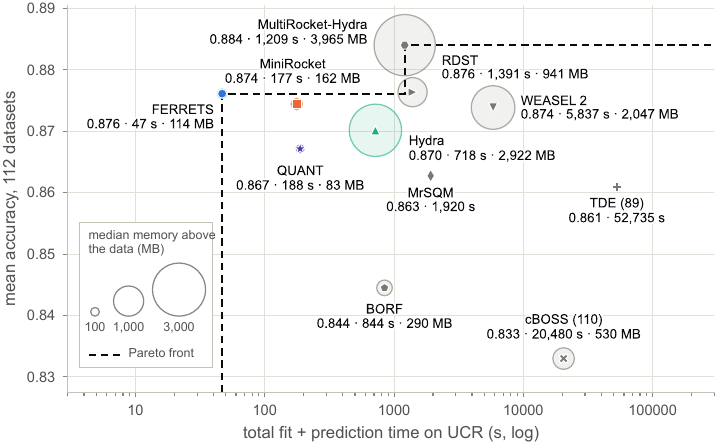}
\caption{Mean accuracy over 30 resamples against fit + prediction time
summed over UCR (resample 10, ten threads where recorded; MultiRocket-Hydra and
RDST~\cite{guillaume2022rdst} in a later session). (n) counts the datasets
completed and the disc area the median memory above the data over
 19 datasets of the memory runs
(Appendix~\ref{app:protocol}), with no disc where it was not measured.}
\label{fig:pareto}
\end{figure}

We present FERRETS (Fast Exact Random Radix Encodings of Time Series), a
dictionary transform that reads no label. Under a linear head, it thus compares
with no training series and holds no candidate word for a selection. Its
skeleton is that of WEASEL~2: random configurations with one dilation each (the
spacing of the samples a window reads), binary words and first differences
(Table~\ref{tab:hyper}). In FERRETS, such a configuration, a
\emph{dictionary}, is one draw of a word source (averages or Fourier coefficients
of the window), a window length and a dilation. Each dictionary turns a window
into a word of eight binary symbols. Read in base two (the radix of the method's
name), the word is an integer below $2^{8}$ and gives its column in the
dictionary's block of 256 columns. The output size is therefore bounded before
any data are seen (Section~\ref{sec:transform}), and two distinct words never
share a column, the sense of ``exact'' in the name. Both word sources normalize
the window and so lose its mean and spread, which a \emph{level word} encodes as
the bins of the window's mean and standard deviation among training quantiles.

Classification is the transform's main application. A ridge classifier on the
counts fits in time linear in the number of training series and
predicts at a constant cost per series (Section~\ref{sec:cost}). On UCR, FERRETS
is the fastest of the eleven methods of Figure~\ref{fig:pareto} under a
ten-thread budget, yet its accuracy is
the highest of the dictionary methods and not significantly different from that
of MiniRocket and QUANT~\cite{dempster2024quant} (quantiles over intervals). On
Multiverse, it is significantly more accurate than both
(Section~\ref{sec:accuracy}). On the MONSTER problems both complete, the
streaming classifier is significantly more accurate than MiniRocket, whose fit
takes 3.7 times as long. FERRETS is also significantly
more accurate than five frozen foundation-model encoders of the MantisV2
report~\cite{feofanov2026mantisv2} on UCR's published split
(Section~\ref{sec:tsfm}).

We make three contributions, two to the transform and one to its use for
classification.
\begin{enumerate}
  \item \emph{A label-free transform whose output size is bounded before any
    data are seen} (Section~\ref{sec:transform}). Its cost is linear in the
    series length. As in the WEASEL~2 paper, a word's column is computed by
    arithmetic, so the transform needs no hash map and a column decodes back to
    its word. A new level word
    gives each window's mean and spread together.
    Removing it lowers accuracy and balanced accuracy on every archive
    (Section~\ref{sec:ablation}).
  \item \emph{Cross-channel words} (Section~\ref{sec:contrib-mv}). On a random
    subset of channels, each channel keeps its own word, and three new words
    describe the subset jointly: a coarse word of all its channels, flags for
    the channels whose word changed since the previous window, and flags for
    those that agree with the first channel. A subset holds at most eight
    channels, so the cost barely grows with their number past eight. While
    these cross channel words only account for a small gain of 0.6
    points of accuracy, we believe the idea is worth iterating on
    (Sections~\ref{sec:cost} and~\ref{sec:ablation}).
  \item \emph{A dictionary classifier that can fit without holding its feature
    matrix} (Section~\ref{sec:contrib-scale}). Since every word's column is
    known before any data are seen, the training counts are encoded once into
    a compressed cache. A
    least-squares head fits in one pass over this cache per epoch for all
    classes together, in 1.7 to 3.3 times
    the batch fit time on a synthetic grid. Two class-balancing rules computed
    from the training set alone, square-root class weights and a shift of the
    decision toward balanced classes, raise balanced accuracy significantly on
    UCR (Sections~\ref{sec:contrib-imbalance} and~\ref{sec:ablation}).
\end{enumerate}

\section{Background}
\label{sec:sota}

\subsection{Definitions}
\label{sec:definitions}

\begin{figure}[tbp]
\centering
\includegraphics[width=\textwidth]{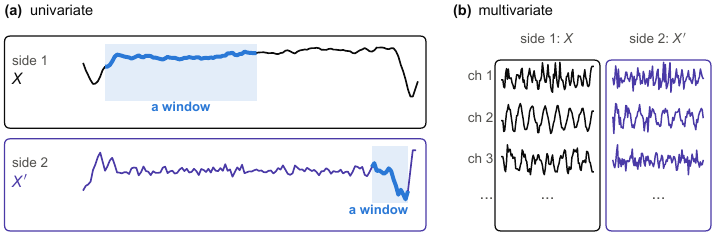}
\caption{The two sides of a series, $X$ (black) and its first difference $X'$
(purple): (a)~an ECG5000 beat with a window in blue on each side; (b)~three of
the six channels of a BasicMotions series.}
\label{fig:def-series}
\end{figure}

\begin{figure}[tbp]
\centering
\includegraphics[width=\textwidth]{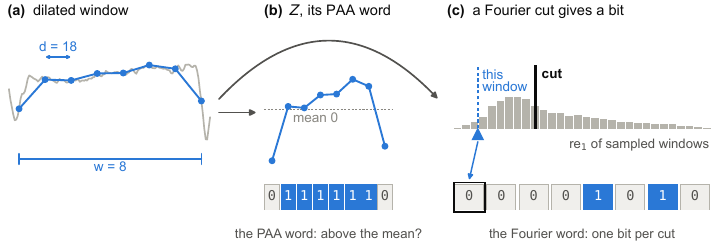}
\caption{(a)~A dilated window ($w=8$, $d=18$, blue) on the series; (b)~the
z-normalized window and its PAA word, one segment per value; (c)~its first Fourier
part (blue) against the cut (black, from the gray cut sample), which gives the
first bit (outlined) of its Fourier word.}
\label{fig:def-word}
\end{figure}

\begin{figure}[tbp]
\centering
\includegraphics[width=\textwidth]{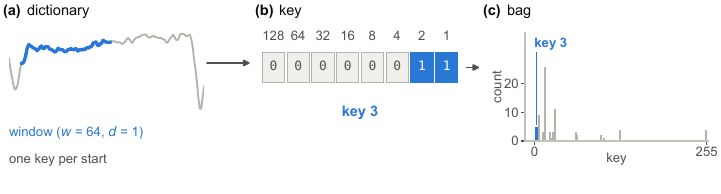}
\caption{One dictionary (PAA, $w=64$, $d=1$) on the series: (a)~a window,
(b)~its word read as a key, (c)~the series' bag over the 256 keys of the block,
the key of (b) in blue.}
\label{fig:def-dictionary}
\end{figure}

\begin{figure}[tbp]
\centering
\includegraphics[width=\textwidth]{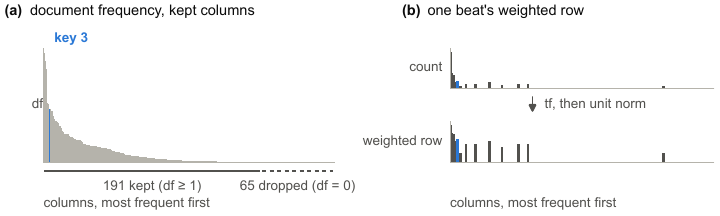}
\caption{One block (PAA, $w=64$, $d=1$): (a)~each key's document frequency over
the training series, kept and dropped columns; (b)~the series' counts and weighted
row over the kept columns, the key of Figure~\ref{fig:def-dictionary} in blue.}
\label{fig:def-row}
\end{figure}

Each definition has its
figure (Figures~\ref{fig:def-series} to~\ref{fig:def-row}), drawn on the ECG5000
beat of Figure~\ref{fig:architecture} (Figure~\ref{fig:def-series}b aside).

\begin{definition}[Series, first difference, side]
\label{def:series}
We consider a time series as a vector $X=(x_1,\ldots,x_m)$ of length $m\ge8$
(the shortest window), and a collection of $n$ series of that length is denoted ${\cal X} = (X_1,\ldots,X_n)$.
For multivariate series, with $C>1$ channels, a series $X$ becomes a matrix
$X\in\mathbb{R}^{C,m}$.

The \emph{first difference} of a series $X$ is denoted $X'$,
with $x'_i = x_{i+1}-x_i$ for $i<m$ and $x'_m = x'_{m-1}$, so that $X'$ also has $m$ values.
For multivariate series, this difference is computed independently per channel.

For all series we consider two \emph{sides}, the raw series and their first differences.
Each side draws its own dictionaries (Definition~\ref{def:dictionary}), independently
of the other side. A dictionary only reads the windows of the side it was drawn on.

\end{definition}

\begin{definition}[Cut, bit, word]
\label{def:word}
A \emph{cut} $\theta \in \mathbb{R}$ turns a value $x$ into the bit $[x>\theta]$,
$1$ if $x>\theta$ and $0$ otherwise. The Fourier and level words take their cuts from
values computed on a \emph{cut sample} of at most 20,000 random training windows of
length $w$ from one side, split evenly over the dilations
(Definition~\ref{def:dictionary}) that side uses with $w$
(for the Fourier word, those of its Fourier dictionaries). The PAA word cuts at $0$.

A \emph{word} is a string of eight bits $u=(u_1,\ldots,u_8)$ computed from a
dilated window $S_{i,w,d}=(x_i,x_{i+d},\ldots,x_{i+(w-1)d})$ of $w$ values
extracted from a series $X$. The window has mean $\mu$ and population standard deviation
$\sigma$, and the z-normalized window is $Z=(S-\mu)/\sigma=(z_0,\ldots,z_{w-1})$.
For numerical stability, a window with $\sigma\le10^{-8}$ is \emph{flat} and has
$Z=0$. The threshold is absolute, so a series of tiny amplitude needs rescaling,
and one far from zero needs shifting.

We define three types of word:
\begin{itemize}
\item The \emph{PAA (Piecewise Aggregate Approximation) word} ($w$ a multiple
of $8$) splits a z-normalized window $Z$ into eight consecutive segments of
$w/8$ values and cuts the mean of each at $\theta=0$, the mean of $Z$,
\begin{equation*}
u_j=\Big[\sum_{t=(j-1)w/8}^{jw/8-1} z_t>0\Big], \qquad j=1,\ldots,8.
\end{equation*}
It is a SAX (Symbolic Aggregate approXimation) word~\cite{SAX} of eight
segments over two symbols (Figure~\ref{fig:window-keys}a), except that SAX gives
ties at $0$ the upper symbol.
\item The \emph{Fourier word} cuts the real and imaginary parts of the first
four Fourier coefficients of $Z$ (the constant term is dropped). For
$h=1,\ldots,4$,
\begin{gather*}
\hat z_h=\frac{1}{w}\sum_{t=0}^{w-1} z_t\,e^{-2\pi\mathrm{i}\,ht/w}, \\
u_{2h-1}=[\operatorname{Re}\hat z_h>\theta_{2h-1}], \qquad
u_{2h}=[\operatorname{Im}\hat z_h>\theta_{2h}],
\end{gather*}
each cut $\theta_j$ being the median of its part over the windows of the cut
sample, plus a tie margin (Appendix~\ref{app:ties}). It is an SFA (Symbolic
Fourier Approximation) word~\cite{SFA} over two symbols. At $w=8$ the last
coefficient is real, so the last bit is always $0$.
\item The \emph{level word} keeps the two moments that the z-normalization
removes. The quantiles $j/16$ of the means of the cut sample's windows
and of their standard deviations give the cuts $\theta^{\mu}_j$ and
$\theta^{\sigma}_j$, $j=1,\ldots,15$, plus tie margins
(Appendix~\ref{app:ties}). The window's mean
falls in bin $b_\mu=\sum_{j}[\mu>\theta^{\mu}_j]$ and its standard deviation in
$b_\sigma$ (a flat window has $b_\sigma = 0$). The level word is the four binary
digits of $b_\mu$ followed by those of $b_\sigma$.
\end{itemize}
\end{definition}

\begin{definition}[Dictionary, key, column, bag]
\label{def:dictionary}
A \emph{dictionary} (WEASEL~2's configuration) is a word source (PAA or Fourier), a
window length $w$ and a dilation $d\ge1$. On its side, it yields the word of every
window $S_{i,w,d}$, $i=1,\ldots,m-(w-1)d$. The \emph{key} of a word $u$
is $u$ read as a binary number, $\kappa(u)=\sum_{j=1}^{8}u_j\,2^{8-j}$, so a dictionary has a
\emph{block} of $2^{8}=256$ keys. The level words of a (side, window, dilation),
of key $16\,b_\mu+b_\sigma$, form one more block. A \emph{column} is one key of one block. Within a block,
the \emph{bag} of a series counts the windows that translate to each key.
\end{definition}

\begin{definition}[Document frequency, kept column, weighted row]
\label{def:tfidf}
A column is \emph{kept} when its key occurs in at least one training bag
(\emph{document frequency} $\mathrm{df}\ge1$); $F$ is the number of kept
columns, over all blocks of both sides, numbered $j=1,\ldots,F$.
For a series $X$, let $\nu_j(X)$ be the number of its windows whose key is the
one of column~$j$. The \emph{sublinear term
frequency}~\cite{manning2008iir} of a count is $\mathrm{tf}(\nu)=1+\ln\nu$
for $\nu\ge1$ and $\mathrm{tf}(0)=0$. The series becomes
\begin{equation*}
\tilde\phi(X)=\big(\mathrm{tf}(\nu_1(X)),\ldots,\mathrm{tf}(\nu_F(X))\big)\in\mathbb{R}^{F},
\qquad
\phi(X)=\frac{\tilde\phi(X)}{\lVert\tilde\phi(X)\rVert_2},
\end{equation*}
its \emph{weighted row}. A key that a
test series takes but no training series did is dropped before scaling. Only a
test series can have no kept key. Its row $\tilde\phi(X)=0$ is then left
unscaled, and a linear head scores it by its intercepts alone.
\end{definition}

\subsection{Related work}
\label{sec:sota-dict}

A \emph{dictionary method} counts symbolic words as a text model counts terms. The
established methods pay at prediction (nearest neighbor) or in memory (a
feature matrix, often selected on the labels). On the other hand, FERRETS keys its words
by arithmetic and selects none. While BOSS VS~\cite{schafer2016bossvs}
also selects no word and scores a test series against one vector per class, it
learns its bins on all the training series and tunes its parameters by
cross-validation over them.

The bag of patterns~\cite{LinLi2009} counts SAX words and classifies by nearest
neighbor. \textbf{BOSS}~\cite{BOSS} does the same
with SFA words in an ensemble, which \textbf{cBOSS}~\cite{cboss} draws at random
and \textbf{TDE}~\cite{tde} (temporal dictionary ensemble) extends with bigrams and a
model-based parameter search. Each member compares a test series with its
training histograms.

\textbf{WEASEL~2}~\cite{schafer2023weasel}, FERRETS' nearest relative, counts
binary SFA words under a ridge classifier.
\textbf{MrSQM}~\cite{mrsqm} selects subsequences of SFA
transforms, and of SAX ones on request, under a logistic regression.
\textbf{BORF}~\cite{borf} counts SAX words on a fixed grid under a ridge classifier.
\textbf{RED
CoMETS}~\cite{redcomets} trains random forests on random SAX and SFA
transformations.

\textbf{Hydra}~\cite{hydra} slides groups of random kernels along the series.
Each kernel sums its response over the positions where it responds most
strongly in its group, and counts the positions where it responds least.
Counting kernels rather than words makes it a
bridge between convolutional and dictionary methods (\href{https://github.com/baraline/ferrets_tsc/blob/main/results/tables/family.csv}{\texttt{family.csv}}).
\textbf{MiniRocket}~\cite{MiniRocket} (dilated convolution kernels under a ridge
classifier) and \textbf{QUANT}~\cite{dempster2024quant} (an interval method) are
not dictionary methods either. With MultiRocket-Hydra~\cite{hydra}
and the ensemble HIVE-COTE~2 (HC2)~\cite{middlehurst2021hc2}, they are the
reference methods (Section~\ref{sec:accuracy}).

For several channels, \textbf{WEASEL+MUSE}~\cite{muse} prefixes every word with
its channel, so its vocabulary grows with $C$. While the papers of MiniRocket and
Hydra describe univariate series only~\cite{MiniRocket,hydra},
their versions in aeon~\cite{aeon} sum a kernel's responses over a few random
channels. Xception~\cite{chollet2017xception} filters each channel on
its own before it mixes them. FERRETS draws a channel subset per dictionary,
computes each channel's word on its own and adds cross-channel words (Section~\ref{sec:contrib-mv}).

\subsubsection{What FERRETS borrows from WEASEL~2, and where it departs.}
WEASEL~2 draws random configurations (a window, a word length, a dilation),
extracts binary SFA words under each from the series and its first difference, and
classifies the counts with a ridge classifier. FERRETS borrows this skeleton
(Tables~\ref{tab:hyper} and~\ref{tab:head}) and departs in three places, the
vocabulary only from its aeon version.

\emph{Words.} WEASEL~2 uses SFA words of up to eight Fourier values chosen by
variance, with equi-depth or equi-width bins drawn per configuration. FERRETS
fixes the values (coefficients $1$ to $4$) and the bins (the median). It adds
the level word and the PAA word. The PAA word is, up to ties, the two-symbol case
of SAX, which MrSQM can pair with SFA.
Prior art keeps a mean or a spread. WEASEL~2 scales each window to unit standard
deviation without subtracting its mean, which keeps the constant Fourier term
among the values it can select~\cite{schafer2023weasel}. BOSS, cBOSS and TDE set
per member whether to subtract the mean~\cite{aeon}.
SAX-SD~\cite{zan2016saxsd} keeps the standard deviation of each segment beside
its mean. The level word adds
the whole window's mean and standard deviation as a word of its own.

\emph{Vocabulary.} In each configuration, aeon's WEASEL~2 keeps by default the
highest-$\chi^2$ words of the series and those of its first difference, up to
$30{,}000/E$ each for an ensemble of $E$ configurations.
It discards words only at its largest ensemble size ($E=150$), and then
at most $56$ of the $256$ values that a binary word of eight symbols takes.
Like the bag of patterns, BOSS, BORF and the WEASEL~2 paper, FERRETS selects no
word on the labels. Its vocabulary is a dense table of $256$ keys per block.
While aeon's WEASEL~2 finds the column of a kept word through a hash map, FERRETS
computes it by arithmetic, as the WEASEL~2 paper does (Section~\ref{sec:transform}).

\emph{Weights and head.} WEASEL~2 passes raw counts to its ridge. FERRETS passes
sublinear term frequencies on rows of unit norm (Definition~\ref{def:tfidf}), the
weighting that BOSS VS~\cite{schafer2016bossvs} gives a test series, its class
vectors adding an inverse document frequency (Section~\ref{sec:ablation}). WEASEL~2 selects the ridge
penalty by the leave-one-out error~\cite{schafer2023weasel,sklearn}. FERRETS does so on small training
sets, without assembling the feature matrix, and holds out data otherwise
(Section~\ref{sec:contrib-scale}).

\section{The FERRETS transform}
\label{sec:contributions}

\begin{figure}[t]
\centering
\includegraphics[width=\textwidth]{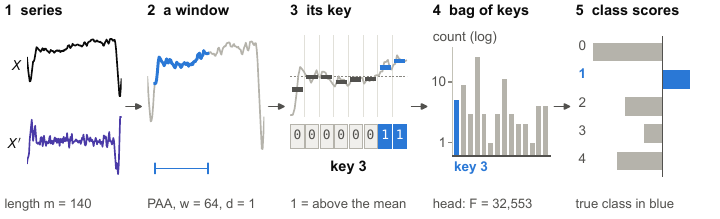}
\caption{FERRETS on one ECG5000 beat. A window of one dictionary becomes an
eight-bit key, the beat's keys fill a bag of counts, and the ridge head scores
the bag.}
\label{fig:architecture}
\end{figure}

The FERRETS transform turns a series into a sparse vector that counts its eight-bit
words, binary PAA and Fourier words and a level word, which bins a window's mean
and standard deviation among training quantiles. A fit draws the dictionaries,
fits their cuts, keys every window and keeps the columns that occur. Transforming a
series then keys its windows and weights its counts. The first four steps of
Figure~\ref{fig:architecture} (series, window, key, bag) are the transform, the last
(scores) its use by a classifier (Section~\ref{sec:classification}).
Table~\ref{tab:hyper} gives the origin of each component.

\subsection{The univariate transform}
\label{sec:transform}

\begin{figure}[tbp]
\centering
\includegraphics[width=\textwidth]{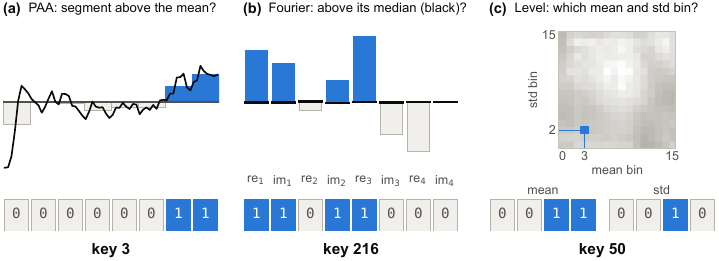}
\caption{One key per window, a 1 in blue: (a)~PAA on ECG5000, (b)~Fourier on
FordA, (c)~the level word of the window of (a), its cell among the training
windows, darker where more fall.}
\label{fig:window-keys}
\end{figure}

Each side, the series or its first difference, draws
$R = 100$ dictionaries, each with a PAA or Fourier source, a window
and a dilation. A window becomes eight bits, and the number they spell picks its
column within its dictionary's block (Figure~\ref{fig:window-keys}). A PAA bit asks
whether the mean of a segment of the window lies above the window's mean, and a
Fourier bit whether the real or imaginary part of a coefficient lies above its
training median.

The transform has four properties:
\begin{itemize}
\item \emph{Label-free.} Every cut is fitted on training windows and every word
  that occurs is kept, so no step reads a label. A fit thus holds no candidate
  word in memory for a selection, and the output does not depend on the task.
\item \emph{Bounded.} Each draw adds at most a word block and a level block of
  $2^{8}$ columns, so the output has at most $2 \cdot R \cdot 2 \cdot 2^{8} =
  102{,}400$ columns over the two sides, whatever the number and the length of
  the series.
\item \emph{Linear.} The keys of a dictionary cost time linear in the length of
  the series (Proposition~\ref{prop:complexity}).
\item \emph{Decodable.} Two different words never share a column, and every column
  decodes to its word and to the dictionary that keyed it, or for a level word
  to its side, window and dilation (Appendix~\ref{app:column-example}). This one-to-one
  map is the sense of ``exact'' in the name, not the search without false
  dismissals of a data-series index such as
  iSAX~\cite{iSAX}.
\end{itemize}

\begin{table}[tbp]
\caption{The configuration of the FERRETS transform. Each component is borrowed
(used as published), adapted (changed as its setting states) or new.}
\label{tab:hyper}
\centering\linespread{1.2}\scriptsize
\setlength{\tabcolsep}{3pt}
\def\rs{\\\addlinespace[4pt]}
\begin{tabular}{@{}>{\raggedright\arraybackslash}p{1.75cm}>{\raggedright\arraybackslash}p{5.2cm}l>{\raggedright\arraybackslash}p{3.45cm}@{}}
\toprule
component & setting & origin & prior work \\
\midrule
\multicolumn{4}{@{}l}{\emph{Transform, one channel (Section~\ref{sec:transform})}} \\
\addlinespace[2pt]
dictionaries & $R = 100$ drawn independently for each side (the series, its first difference), seed 0, duplicates merged
  & adapted & WEASEL~2~\cite{schafer2023weasel} \rs
source & PAA or Fourier, equal probability
  & adapted & SAX~\cite{SAX}, SFA~\cite{SFA}, paired in MrSQM~\cite{mrsqm} \rs
window $w$ & uniform on $\{8,16,24,32,48,64\}$, $w \le m$
  & adapted & WEASEL~2 \rs
dilation $d$ & $\lfloor 2^{U} \rfloor$, $U \sim \mathcal{U}[0, \log_2(d_{\max}+1))$, $d_{\max} = \lfloor (m-1)/(w-1) \rfloor$, the $+1$ letting $d_{\max}$ be drawn
  & adapted & ROCKET's draw~\cite{rocket}, as in WEASEL~2 \rs
PAA word & 8 segments of $w/8$ values, bit 1 where the segment mean exceeds the window mean, 0 on a tie
  & adapted & SAX at two symbols~\cite{SAX} \rs
Fourier word & coefficients 1--4 (real, imaginary) over $w\sigma$, bit 1 above the median of the family's cut sample ($\le 20{,}000$ windows)
  & adapted & SFA at two symbols~\cite{SFA} \rs
level word & mean and standard deviation, 4 bits each (bins among 15 quantiles of the cut sample of a side and $w$), a block per (side, $w$, $d$)
  & new & SFA's equi-depth bins~\cite{SFA}, a mean over the spread kept in WEASEL~2, a spread in SAX-SD~\cite{zan2016saxsd} \rs
key, layout & the word read in base two, $2^{8}$ columns per block, $F \le 102{,}400$, a flat window keyed as if $Z=0$
  & adapted & iSAX~\cite{iSAX}, aeon's SFA~\cite{aeon}, WEASEL~2's blocks \rs
columns & every column present in a training series kept, every window counted
  & borrowed & BORF~\cite{borf} \rs
weighting & sublinear tf and unit $\ell_2$ rows for every series, BOSS VS's weighting of a test series (Definition~\ref{def:tfidf})
  & adapted & BOSS VS~\cite{schafer2016bossvs} \rs
off by default & idf, numerosity reduction (a run of equal consecutive words counts once), pruning of the columns of fewer than 2 training series (Section~\ref{sec:ablation})
  & borrowed & scikit-learn~\cite{sklearn}, SAX~\cite{SAX} \\
\midrule
\multicolumn{4}{@{}l}{\emph{Transform, $C \ge 2$ channels (Section~\ref{sec:contrib-mv})}} \\
\addlinespace[2pt]
sides & the channels, their signed first differences
  & adapted & WEASEL+MUSE~\cite{muse} \rs
families & $R$ draws per side, each on $g \in \{2,4,8\}$ channels ($g \le C$, without replacement), equal (source, window, channels) forming a family
  & adapted & multivariate MiniRocket and Hydra in aeon~\cite{aeon} \rs
blocks & a block of $2^{8}$ columns per word of a (family, dilation), a level block per (side, channels, $w$, $d$) with $8/g$ bits per channel, $F \le 614{,}400$
  & adapted & WEASEL+MUSE's channel blocks~\cite{muse} \rs
cross-channel words & coarse word, change flags, agreement flags (Definition~\ref{def:family-words})
  & new & mixing after a per-channel step, as in Xception~\cite{chollet2017xception} \\
\bottomrule
\end{tabular}
\end{table}

\paragraph{Draws.} A dictionary of window $w$ draws its dilation $d$ log-uniformly
up to the largest that its window admits, $\lfloor (m-1)/(w-1) \rfloor$
(Table~\ref{tab:hyper}), so over the draws the window span $(w-1)\,d+1$ spreads
about log-uniformly from $w$ to about $m$. Equal draws merge, so a side keeps at
most $R$ dictionaries, fewer on a short series. Dictionaries of one side that differ only in
dilation form a \emph{family}. A Fourier family fits its cuts once, on a cut
sample (Definition~\ref{def:word}) spread over its dilations
(Appendix~\ref{app:ties}).

\paragraph{Level word.} The PAA and Fourier words read a z-normalized window, so
a window $S$ and its copy $a_1 S + a_0$ get the same word, whatever the scale
$a_1 > 0$ and the offset $a_0$, unless one of the two is flat
(Definition~\ref{def:word}). Classes that differ only by the level or the
amplitude of their windows thus get the same words. The level word gives both
back, at sixteen levels each (Figure~\ref{fig:window-keys}c). A window beyond the
cuts falls in the first or last bin, however far it lies outside the training range.

\paragraph{Weighting.} Each count $\nu$ becomes $1 + \ln\nu$, which damps the
words a series repeats, and each row is scaled to unit norm (Definition~\ref{def:tfidf}), so that a series that
repeats a few words and one that spreads over many reach the head at the same norm. Three bag transforms are
off by default: inverse document frequency, numerosity reduction and pruning
(Table~\ref{tab:hyper}, Section~\ref{sec:ablation}).

\paragraph{Output space.} FERRETS lays the blocks of all dictionaries end to end,
as WEASEL~2 does~\cite{schafer2023weasel}. A word's column is the first column
of its block plus its key $\kappa(u)$, so it is computed from the draw, with no
hash map (Appendix~\ref{app:column-example}). The fitted transform stores no
word string, only its cuts and integer tables built at fit.

The kept vocabulary is smaller than the bound, since a dictionary of large
dilation admits few windows and keys that no training window produces are not
kept. On UCR it stays at a median of 0.44 of the bound and at most
0.75 of it.

\begin{proposition}[Cost of the keys]
\label{prop:complexity}
Let a dictionary have a window $w \le 64$ and any dilation.
The keys of all windows of a series of length $m$ are computed in $O(m)$ time.
With $V$ the number of distinct dictionaries left once equal draws collapse,
keying $n$ series under them and their level blocks takes $O(n\,V\,m)$ time.
\end{proposition}

\noindent The proof, in Appendix~\ref{app:keys-proof}, reads the samples one
dilation apart as consecutive values, as WEASEL~2 does~\cite{schafer2023weasel},
and takes each window's moments and Fourier coefficients from prefix sums
(running totals) along them, as the UCR Suite~\cite{UCRSuite} does for the moments.

\paragraph{From values to bits.} The exactness of the keys covers the step from bits
to columns, not the step from values to bits, which compares floating-point numbers.
On UCR, the univariate kernel's PAA and level keys differ from the definition
evaluated in exact arithmetic on at most 1.0\,\% of a dataset's
sampled training windows, and identical windows share a key in every case checked
(Appendix~\ref{app:ties}). The Fourier keys, whose parts involve sines and
cosines, enter only the check on identical windows. The
multivariate keying, on whose ties the flags rely, is untested.

\paragraph{Fixed settings.} The window set and the tie margins were set once and
not varied. Windows span 8 to 64 values, longer structure entering
through dilation. Windows scaled to the series length were not tried. Words keep
the eight binary symbols of WEASEL~2, and each further bit would double the
vocabulary. The segments, coefficients and cuts
that fill the eight bits (Table~\ref{tab:hyper}) were not varied either. Only
shorter words were compared (Section~\ref{sec:ablation}).

\subsection{Multivariate input: channel-wise and cross-channel words}
\label{sec:contrib-mv}

\begin{figure}[!htb]
\centering
\includegraphics[width=\textwidth]{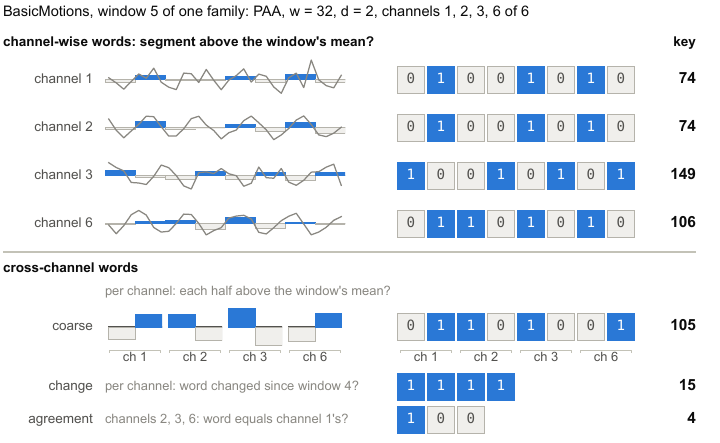}
\caption{Window~5 of one BasicMotions family, with each channel's window and
channel-wise word, then the cross-channel words (level block not drawn; the
half means enlarged). Window~4 is one dilation earlier. A blue cell is a 1.}
\label{fig:mv-words}
\end{figure}

WEASEL+MUSE keys each channel on its own~\cite{muse}, which multiplies the
vocabulary by the number of channels $C$ and relates no two channels. A linear
head sums the weights of the columns it reads, so it counts a pattern on one
channel the same whatever the others do. FERRETS keys a random subset of at
most eight channels together instead, the \emph{channel layout}. Each channel
gets its own word, and three more words say what the subset does jointly
(Figure~\ref{fig:mv-words}). From eight channels on, the bounds on the keying
work and on the vocabulary no longer grow with $C$.
\begin{itemize}
\item \emph{Families on subsets.} Each side, the channels or their first
  differences (Definition~\ref{def:series}), draws $R$ dictionaries as in
  Section~\ref{sec:transform}. Each reads $g$ channels chosen without
  replacement, $g$ uniform over the members of $\{2, 4, 8\}$ not above $C$, and
  draws with the same source, window and channels form one family. A power of
  two gives each channel an equal share, $8/g$ bits, of the coarse and level
  words.
\item \emph{Channel-wise words.} As in WEASEL+MUSE, each channel's word has a
  block of its own.
\item \emph{Cross-channel words.} The three words relate the channels after the
  per-channel step. The \emph{coarse word}
  records the shapes the channels take together. One of its columns counts, for
  example, the windows where the first channel rises while the second falls.
  The price is resolution, since each channel gives this eight-bit key a
  shortened word of $8/g$ bits. The \emph{change flags} record which channels
  change shape at the same time. One of their columns counts the windows where
  the first and third channels changed their word since the window one dilation
  earlier, and the others did not. The \emph{agreement flags} record which
  channels share a shape. One of their columns counts the windows where the
  second channel has the first channel's word, whatever that word is, and the
  others do not.
\end{itemize}

\begin{definition}[Words of a family]
\label{def:family-words}
Let a family have $g \ge 2$ channels, numbered $r = 1, \ldots, g$ by increasing
index. At a dilation $d$ and a start $i$, channel $r$ has the window $S^{(r)}_i$
(Definition~\ref{def:word}) with mean $\mu^{(r)}_i$ and standard deviation
$\sigma^{(r)}_i$. Its \emph{channel-wise word} is $u^{(r)}_i = u(S^{(r)}_i)$, the
word of its window under the family's source, a Fourier word taking its cuts
from channel~$r$'s windows at the family's dilations. The family has four more words.
Each is a string of bits, channel~$1$ first, read as a binary number
(Definition~\ref{def:dictionary}).
\begin{itemize}
\item The \emph{coarse word} joins one short word of $8/g$ bits per channel. In
  a Fourier family, channel $r$ gives the first $8/g$ bits of $u^{(r)}_i$, its
  lowest frequencies, down to the real part of coefficient~$1$ alone at $g = 8$.
  In a PAA family, channel $r$ gives the PAA word of
  $S^{(r)}_i$ over $8/g$ segments instead of eight, one bit per quarter of the
  window at $g = 2$ and per half at $g = 4$. At $g = 8$, where one segment would
  always give $0$, it gives the first bit of the word over two halves.
\item The \emph{change flags} are the $g$ bits $[u^{(r)}_i \ne u^{(r)}_{i-d}]$,
  each $0$ when $i \le d$.
\item The \emph{agreement flags} are the $g-1$ bits $[u^{(r)}_i = u^{(1)}_i]$,
  $r = 2, \ldots, g$.
\item The \emph{level word} takes, per channel, the first $\lceil 4/g \rceil$
  binary digits of the bin of $\mu^{(r)}_i$, then the first $\lfloor 4/g \rfloor$
  of the bin of $\sigma^{(r)}_i$, both among cuts drawn from channel~$r$ alone
  (Definition~\ref{def:word}).
\end{itemize}
\end{definition}

\noindent The agreement flags compare with the subset's lowest-index channel,
so the words depend on the order of the channels, and the effect of another order
on accuracy was not measured. The multivariate level word is coarser than the
univariate one, down to a single mean bit per channel at $g = 8$.

\paragraph{Layout and bound.} A family owns $g + 3$ blocks per dilation, one per
channel-wise or cross-channel word, and a level block when it holds one. Columns
follow the arithmetic of Section~\ref{sec:transform}. With $g_{\max}$ the
largest of $2, 4, 8$ not above $C$, keying $n$ series costs
$O(n\,R\,g_{\max}\,m)$ per side. The transform has at most
$2R\,(g_{\max} + 4)\,2^{8}$ columns, 307{,}200 for $2 \le C \le 3$, 409{,}600 for
$4 \le C \le 7$ and 614{,}400 from $C = 8$ on.

\paragraph{Coverage.} The price of the cap at eight channels is that a channel
can go unread. A subset of $g$ channels misses a given channel with probability
$1 - g/C$, and from $C = 8$ on $g$ is $2$, $4$ or $8$ with equal probability,
$14/3$ on average, so the $R$ independent subsets of a side all miss it with
probability
\begin{equation*}
\Big(1 - \frac{14}{3C}\Big)^{R} \approx e^{-14R/(3C)}.
\end{equation*}
At $C = 1{,}345$, the largest channel count in Multiverse, this probability is
0.71, and neither side reads the channel with probability 0.50. With seed~0,
every run on a problem leaves the same channels unread. The per-channel layout, which keys every channel
on its own as WEASEL+MUSE does, reads them all, and Section~\ref{sec:ablation}
compares the two.

\section{Classification with FERRETS}
\label{sec:classification}

Following MiniRocket~\cite{MiniRocket}, FERRETS classifies its weighted rows, the
bags of Definition~\ref{def:tfidf} scaled to unit norm, with a linear head. It has
two heads, a ridge head as in MiniRocket (the batch head) and, where the training
matrix outgrows memory, a streaming head (Section~\ref{sec:contrib-scale}). Each
head keeps one configuration for all datasets (Table~\ref{tab:head}). The only
settings chosen per dataset are the penalty $\alpha$ and the prior-shift weight
$\lambda$ of the ridge head (Section~\ref{sec:contrib-imbalance}), both from its
training data. Their grids were set once and never varied, while the step and the
number of epochs of the streaming head were chosen on the UCR development third
(Appendix~\ref{app:protocol}).

\paragraph{Head.} The training collection $\mathcal{X} = (X_1, \ldots, X_n)$
(Definition~\ref{def:series}) carries the labels $y_1, \ldots, y_n$ of $K$
classes, and $\Phi \in \mathbb{R}^{n,F}$ stacks its weighted rows, row $s$ being
$\phi(X_s)$. Class $c$ scores the weighted row of a series as
$f_c = \beta_c^{\top}\phi + \beta_{c,0}$, and the head predicts $\arg\max_c f_c$
(on two classes, the sign of one score). The ridge head fits each score by least
squares to the one-vs-rest targets $\tau_{sc}$, $+1$ if $y_s = c$ and $-1$
otherwise, with the penalty $\alpha\lVert\beta_c\rVert_2^{2}$ and an unpenalized
intercept $\beta_{c,0}$. Each squared error is weighted by the class weight
$\gamma_{y_s}$ of its series (Section~\ref{sec:contrib-imbalance}).

\paragraph{Prediction.} Prediction builds no feature matrix, and classifying a
series adds one sparse dot product per class to the cost of transforming it. Each
score is the
dot product of the coefficients with the row before its $\ell_2$ scaling
(Definition~\ref{def:tfidf}), divided by the norm of that row, plus the intercept.
The dot products and the squared norm are all sums over the nonzero columns of the
series, so the kernel accumulates them while it reads the bag. No step compares
two series or counts the candidate words of the training set to select some of
them (Section~\ref{sec:sota-dict}). A test series thus costs its keys and its
scores, whatever the size of the training set.

\begin{table}[!htb]
\caption{The configuration of the heads. Each component is borrowed (used as
published), adapted (changed as its setting states) or new.}
\label{tab:head}
\centering\linespread{1.2}\scriptsize
\setlength{\tabcolsep}{3pt}
\def\rs{\\\addlinespace[4pt]}
\begin{tabular}{@{}>{\raggedright\arraybackslash}p{1.75cm}>{\raggedright\arraybackslash}p{5.2cm}l>{\raggedright\arraybackslash}p{3.45cm}@{}}
\toprule
component & setting & origin & prior work \\
\midrule
ridge head & $\pm 1$ one-vs-rest targets, intercept by centering, $\alpha \in \{10^{-3}, 10^{-2}, 10^{-1}, 1, 10\}$
  & adapted & MiniRocket~\cite{MiniRocket} \rs
class weights & $\gamma_c \propto 1/\sqrt{n_c}$ (mean 1) in the ridge loss, $\omega_c = \sqrt{n_{\min}/n_c}$ on streaming updates, $n_c$ the size of class $c$, $n_{\min}$ that of the smallest (Section~\ref{sec:contrib-imbalance})
  & adapted & square-root sampling~\cite{mahajan2018weakly} \rs
prior shift & ridge heads predict $\arg\max_c (f_c - \lambda \bar y_c)$ (Section~\ref{sec:contrib-imbalance}), $\lambda \in \{0, .25, .5, .75, 1\}$ by the one-standard-error rule on out-of-sample decisions
  & adapted & logit adjustment~\cite{menon2021logit}, one-standard-error rule~\cite{Breiman1984} \rs
dual route & $n \le 3{,}000$, $\alpha$ by exact class-weighted leave-one-out squared error (Section~\ref{sec:contrib-scale})
  & borrowed & scikit-learn's leave-one-out ridge~\cite{sklearn}, one eigendecomposition for all $\alpha$~\cite{rifkin2007notes} \rs
iterative route & $n > 3{,}000$, LSQR with tolerance $10^{-2}$ and $\le 200$ steps, $\alpha$ by accuracy on a 25\,\% hold-out, refit on all rows
  & adapted & LSQR~\cite{paige1982lsqr}, one bidiagonalization for all $\alpha$~\cite{oleary1981bidiagonalization} \rs
streaming head & least squares, class-weighted (Equation~\ref{eq:stream}), step 1, $\ell_2$ penalty $10^{-5}$ per series, 20 epochs, chunks of $2.5 \times 10^{7}$ nonzeros
  & adapted & averaged SGD at a constant step~\cite{BachMoulines2013}, scikit-learn's SGD~\cite{sklearn} \\
\bottomrule
\end{tabular}
\end{table}

\subsection{Class imbalance}
\label{sec:contrib-imbalance}

A least-squares head leans toward the larger classes. Each score carries the prior
of its class as its intercept, and the penalty pulls the score toward that prior. FERRETS corrects this bias with two
rules computed from the training set alone, neither of which changes a prediction
when the classes are exactly balanced: class weights in the loss of both heads
and, for the batch head, a shift of each score that removes part of its prior. On
UCR, the rules raise balanced accuracy significantly, while accuracy is not
significantly different (Section~\ref{sec:ablation}).

\paragraph{Class weights.} The batch head weights the squared error of a training
series of class $c$ by $\gamma_c \propto 1/\sqrt{n_c}$, $n_c$ being the size of
the class. The weights are scaled to mean 1 over the training series, so the
penalty grid keeps the scale it has without weights. Up to a constant, these are
the square roots of the \emph{balanced} weights of scikit-learn~\cite{sklearn}, a
compromise between no weighting and full balancing. In expectation, they weight
the loss as drawing class $c$ with probability $\propto \sqrt{n_c}$ does, the plain
form of the square-root sampling of Mahajan et~al.~\cite{mahajan2018weakly}. On
the shipped model, they were compared with no weights only
(Section~\ref{sec:ablation}), not with full balancing or other exponents. The
streaming head multiplies each update by $\omega_c = \sqrt{n_{\min}/n_c}$,
$n_{\min}$ being the size of the smallest class. These are the same weights
divided by their maximum, so no class takes a step above the constant step of 1.
Weights of mean 1 can give a small class a step above 2, at which the updates
diverge on rows of unit norm.

\paragraph{Prior shift.} Let $\pi_c \propto \sqrt{n_c}$ be the weighted share of
class $c$ and $\bar y_c = 2\pi_c - 1$ the weighted mean of its $\pm 1$ target.
With an unpenalized intercept, the ridge head scores a series $X$ by
\begin{equation*}
f_c(X) = \bar y_c + \beta_c^{\top}\big(\phi(X) - \bar\phi\big),
\end{equation*}
$\bar\phi$ being the weighted mean of the features. The first term is the prior
of the class and the second the learned part. As the penalty grows, the learned
part goes to zero and every score to its prior $\bar y_c$, so the head predicts
the majority class. FERRETS removes a share $\lambda$ of the prior from each score
and predicts
\begin{equation*}
\hat y(X) = \arg\max_{c}\,\big(f_c(X) - \lambda\,\bar y_c\big),
\qquad \lambda \in \{0,\, 0.25,\, 0.5,\, 0.75,\, 1\},
\end{equation*}
so that $\lambda = 0$ is the plain head and $\lambda = 1$ keeps the learned part
alone. On two classes, the single score is shifted the same way. Note that the
shift moves only the intercepts, so it adds nothing to the cost of prediction. It
is an analog, for least-squares scores, of the logit adjustment of Menon
et~al.~\cite{menon2021logit}, which subtracts a multiple of the log class prior
from each logit, the multiple playing the part of $\lambda$.

\paragraph{Choosing the shift.} The shift is chosen per dataset on out-of-sample
decisions that both batch routes already produce for the training series: the
exact leave-one-out decisions of the dual route and the hold-out decisions of the
iterative route (Section~\ref{sec:contrib-scale}). Each $\lambda$ is scored by
the mean of its accuracy, balanced accuracy and macro F1. FERRETS takes the
smallest $\lambda$ within one binomial standard error of the best score, so the plain head
stands unless a shift beats it by more than that margin
(Appendix~\ref{app:lambda}). The streaming head is not shifted, since it produces
no such decisions, and holding out part of its cache would either remove series
from its fit or require a second fit. It corrects the imbalance through its class
weights alone.

\subsection{Fitting at scale}
\label{sec:contrib-scale}

\begin{figure}[!htb]
\centering
\includegraphics[width=\textwidth]{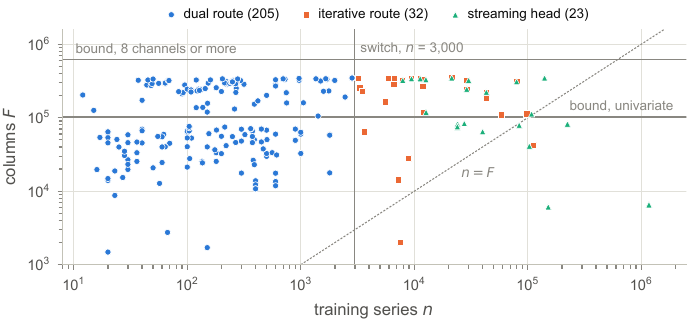}
\caption{The problems FERRETS completes, by fit regime (batch head on UCR and
Multiverse, streaming head on MONSTER). Hairlines: the switch between the batch
routes, the bounds on $F$ and $n = F$.}
\label{fig:fit-regimes}
\end{figure}

The fit has three regimes, which trade exactness for scale:
\begin{itemize}
\item the batch \emph{dual route}, up to 3{,}000 training
series, solves the ridge in closed form on the Gram matrix,
\item the batch \emph{iterative route}, beyond, solves each class by LSQR, an
iterative least-squares solver, and keeps the feature matrix $\Phi$ in memory,
\item the \emph{streaming head} never holds $\Phi$, only an encoded cache of the
counts, in memory or on disk.
\end{itemize}
For $K$ classes, a step of the iterative route and an epoch of the streaming head
take time linear in $KN$, with $N \le nF$ the number of nonzero entries of $\Phi$.
The iterative route also keeps the basis that LSQR builds for each class
orthogonal, which adds time proportional to $kKF$ at step $k$. Under the cap of
200 steps, this cost stays linear in $KN$, since $F \le N$. Neither the cap nor
the 20 epochs grows with $n$. Given that the
bound on $F$ does not depend on $n$, these two regimes take time at most linear in
$n$ for a given number of classes. In the experiments, $F$
stays under its bound while the training set grows from the smallest Multiverse
problem to the largest that FERRETS completes at MONSTER
(Figure~\ref{fig:fit-regimes}). Section~\ref{sec:cost} measures these costs.

\paragraph{Dual route.} The Gram matrix $G = \Phi\Phi^{\top}$ has one row and one
column per training series, so a system of size $n$ replaces one of size $F$, the
smaller of the two on each of the 205 problems the route fits
(Figure~\ref{fig:fit-regimes}). It is summed over slabs of consecutive columns,
each made dense within a fixed memory budget, in $O(n^{2}F)$ time. One
eigendecomposition of the class-weighted Gram matrix of the centered rows gives
the leave-one-out error of every penalty in closed form~\cite{rifkin2007notes}.
The solution, computed in double precision from the rows as stored in single
precision, is exact up to rounding (Appendix~\ref{app:ridge-routes}). But the eigendecomposition takes $O(n^{3})$
time, which leaves larger training sets to the iterative route. The switch bounds
the Gram matrix at 9{,}000{,}000 entries. It is a fixed setting of the model, not
the point where the two routes take the same time.

\paragraph{Iterative route.} Each class is solved by LSQR~\cite{paige1982lsqr},
whose basis does not depend on the penalty. The route stores this basis and keeps
it orthogonal, so one run per class serves the whole
grid~\cite{oleary1981bidiagonalization}. Each run stops
at a loose tolerance, which shortens the fit, or at a cap on its steps. The
penalty kept is the one most accurate on a hold-out (Table~\ref{tab:head}). The
route is thus an approximate ridge. On the 17 problems of 3{,}000 to
10{,}000 series where the exact ridge also runs, its accuracy is not significantly
different from that of the exact ridge, 0.4 points lower on average
(Appendix~\ref{app:ridge-routes}). Above that size, the only check left is the
batch head against the streaming head on the nine MONSTER problems
(Section~\ref{sec:cost}).

\paragraph{Streaming head.} The streaming fit transforms each training series
once, in chunks and in a fixed random order. It appends their counts to the cache
(Appendix~\ref{app:protocol}), in the byte-aligned gap code that inverted files use
for postings~\cite{zobel2006inverted}, where a row stores the differences between
its successive column ids. On the cache, a pass over $\Phi$ means reading and
decoding all of it. The iterative route would need many such passes
(Appendix~\ref{app:stream-objective}), while the streaming head makes one pass per
epoch for all classes together. The head runs averaged stochastic gradient
descent~\cite{BachMoulines2013}, as in scikit-learn~\cite{sklearn}, the goal being
a rough optimum reached in few passes. Each score $c$ is fitted with a constant
step of 1 on the objective
\begin{equation}
\label{eq:stream}
J_c(\beta_c, \beta_{c,0}) = \frac{1}{n}\sum_{s=1}^{n}
\frac{\omega_{y_s}}{2}\big(\tau_{sc} - \beta_c^{\top}\phi(X_s) -
\beta_{c,0}\big)^{2} + \frac{10^{-5}}{2}\,\lVert\beta_c\rVert_2^{2},
\end{equation}
with the targets $\tau_{sc}$ of Section~\ref{sec:classification} and the class
weights $\omega$ of Section~\ref{sec:contrib-imbalance}. On UCR, the streaming
head is significantly less accurate than the batch head
(Section~\ref{sec:ablation}), a trade-off between accuracy and memory.

\paragraph{Threads.} Each stage splits its work so that no two threads write the
same data. The parts over which a row's dot products are summed depend on the
row's length, never on the thread count, so the keys and the streaming head's
coefficients are the same at every thread count.

\section{Experiments}
\label{sec:experiments}

The experiments ask where FERRETS is worth using and where it is not, cost being
its main strength. Each experiment evaluates the transform through the classifier
of Section~\ref{sec:classification}, against each competitor's whole pipeline.
All our results files, such as \href{https://github.com/baraline/ferrets_tsc/blob/main/results/tables/overall.csv}{\texttt{overall.csv}}, come from the repository
\url{https://github.com/baraline/ferrets_tsc}, which also holds the published
results as we read them.

\subsection{Protocol}
\label{sec:protocol}

\newcommand{\raggedcell}{\raggedright\let\\\tabularnewline}

Table~\ref{tab:archives} lists the three archives that carry the comparison, a
method's score being its mean over the archive's splits. A UCR dataset is scored
on its \emph{published split} and on resamples that re-split the pooled series,
keeping the class proportions and set sizes~\cite{bakeoffredux}. Multiverse
uses the archive's own equal-length, complete versions, except the unequal-length
InsectWingbeat, and MONSTER the raw arrays of its Hugging Face hub release. The
Multiverse core is that of the archive~\cite{multiverse} less
BenzeneConcentration\_disc, one of whose channels is a function of the label in
the version the results ran on.
UCR and Multiverse also assign each problem an application
domain~\cite{UCRArchive,multiverse}.

\begin{table}[!htb]
\caption{The three archives: size, splits and the origin of the competitors'
results. Methods with published UCR results are also re-run for time and memory
(Section~\ref{sec:cost}).}
\label{tab:archives}
\centering\footnotesize
\begin{tabular}{@{}p{1.65cm}p{2.9cm}p{2.4cm}p{1.8cm}p{2.5cm}@{}}
\toprule
 & \raggedcell problems & \raggedcell splits & \raggedcell published results
 & \raggedcell run here for accuracy \\
\midrule
\raggedcell UCR
 & \raggedcell 112 univariate datasets~\cite{UCRArchive}
 & \raggedcell 30 resamples, the first the published split
 & \raggedcell 19 methods
 & \raggedcell FERRETS, BORF \\
\addlinespace
\raggedcell Multiverse
 & \raggedcell 133 multivariate problems, 65 in the core~\cite{multiverse}
 & \raggedcell one run on the published split
 & \raggedcell 31 methods
 & \raggedcell FERRETS, MiniRocket, Hydra, BORF \\
\addlinespace
\raggedcell MONSTER
 & \raggedcell 29 problems, 10 univariate and 19 multivariate~\cite{MONSTER}
 & \raggedcell five predefined cross-validation folds, some grouped by subject or location
 & \raggedcell seven methods~\cite{monsterresults}
 & \raggedcell FERRETS (streaming), MiniRocket \\
\bottomrule
\end{tabular}
\end{table}

We run FERRETS on every problem that the gate and the data allow, and a
competitor wherever its archive publishes no result for it. A feasibility gate
skips a cell whose predicted time or memory exceeds the budget, and a disk rule
one whose cache would not fit the disk. The competitors keep their defaults,
except WEASEL+MUSE's logistic head and a training-time limit for TDE and cBOSS on
the size grid (Appendix~\ref{app:protocol}). Every run gets
ten threads and one hour per fold, data loading included, with the exceptions
that the same appendix lists.

We pair the comparisons by dataset~\cite{Demsar2006}, on the datasets that every
compared method completes, and test them with two-sided Wilcoxon signed-rank
tests~\cite{Wilcoxon1945} under Holm's correction~\cite{Holm1979}. We call a
difference significant when its corrected $p$ is below the 0.05 level. The primary
comparison, which Figure~\ref{fig:overall-wins} draws, sets FERRETS against
MiniRocket and QUANT, the competitors closest in fit and prediction time on UCR,
with the correction made within each archive. The file \href{https://github.com/baraline/ferrets_tsc/blob/main/results/tables/overall.csv}{\texttt{overall.csv}} adds
percentile bootstrap intervals of the paired mean difference from 10{,}000 draws.
The multi-comparison matrices~\cite{mcm} of
Figures~\ref{fig:mcm-ucr} to~\ref{fig:mcm-monster} add, for each family, the
method most accurate on UCR among those with results on UCR and Multiverse:
MultiRocket-Hydra, RDST, HC2, LITETime~\cite{LITE} and FreshPRINCE. They test each
pair on the problems both methods complete and, unlike the original matrices,
correct over all their pairs.

FERRETS' defaults were chosen on a development third of each archive,
37 UCR datasets, 44 Multiverse problems and
9 MONSTER problems, drawn at random within three bands of
series length. Appendix~\ref{app:protocol} lists the
cells behind each choice, and every other problem is outside the development third
at every split. FERRETS runs with seed~0. Neither another seed on UCR nor averaging
the Multiverse problems that come from one source changes a primary verdict
(Appendix~\ref{app:protocol}).

\subsection{Scalability: time, memory and model size}
\label{sec:cost}

\begin{figure}[tb]
\centering
\includegraphics[width=\textwidth]{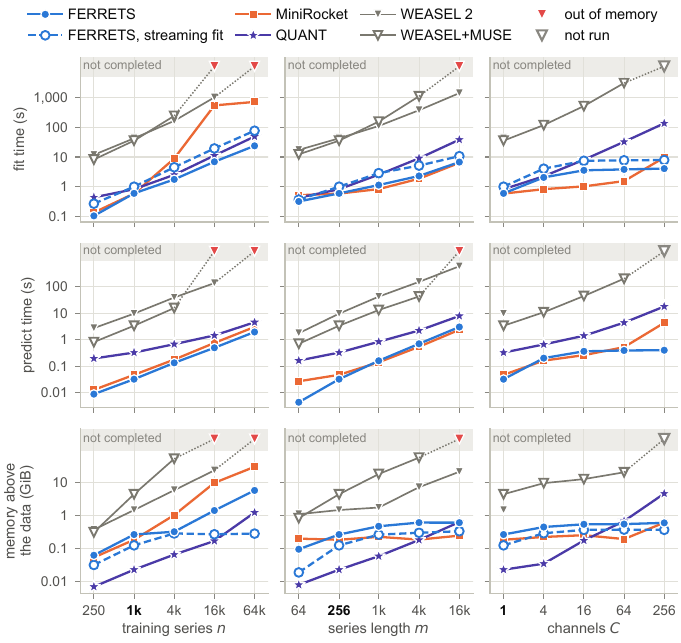}
\caption{Fit and predict time and memory, one axis at a time around
$n=1{,}000$, $m=256$, $C=1$ (bold ticks), medians of three repeats of one draw of
the dictionaries (one repeat for WEASEL+MUSE at $C=64$). Both FERRETS fits
predict alike.}
\label{fig:cost-grid}
\end{figure}

On a synthetic grid that moves one axis at a time, FERRETS' fit time grows
linearly with the number $n$ of training series once the batch head turns
iterative, above 3{,}000, and sublinearly with the series length
(Figure~\ref{fig:cost-grid}, \href{https://github.com/baraline/ferrets_tsc/blob/main/results/tables/cost-scaling.csv}{\texttt{cost-scaling.csv}}). A test series costs the
same whatever $n$. Past eight channels the cost barely grows, since a family
keys at most eight channels, and at the largest grid point MiniRocket takes
2.4 times as long to fit. The price of this bound is that a
channel can go unread.

FERRETS is far cheaper than the dictionary methods it descends from. WEASEL~2
and WEASEL+MUSE take 39 to 837 times
as long to fit on the grid points they complete, and Figure~\ref{fig:cost-grid}
marks where they run out of memory on larger ones that FERRETS completes. The gap
does not come from the threads, since at one thread WEASEL~2 still takes
21 times FERRETS' fit time.

In Figure~\ref{fig:pareto}, no method is both more accurate and faster than
FERRETS on UCR. MiniRocket and QUANT, not significantly different in accuracy,
take 3.78 and 4.01 times as long to fit and
predict (\href{https://github.com/baraline/ferrets_tsc/blob/main/results/tables/cost-ucr.csv}{\texttt{cost-ucr.csv}}), and the two methods with a higher mean
accuracy, MultiRocket-Hydra and RDST, take 24.7 and
28.4 times as long. MiniRocket's longer fit comes from the
four datasets with more than 3{,}000 training series, where it
takes 15 times as long. There, aeon fits the leave-one-out ridge
of scikit-learn~\cite{sklearn} at a cost cubic in $n$, while FERRETS hands over to
LSQR. On the other hand, FERRETS needs its threads on small data, and at one
thread on 19 datasets of the memory runs
(Appendix~\ref{app:protocol}) it takes 1.6 times as long as
MiniRocket to fit.

On Multiverse, FERRETS is more accurate than MiniRocket and faster, at
0.24 of its fit time over the 102 problems every method
completes, but it needs more memory and larger models (\href{https://github.com/baraline/ferrets_tsc/blob/main/results/tables/cost-mv.csv}{\texttt{cost-mv.csv}}).
MiniRocket stays cheaper on the smaller problems, and on the grid it fits faster
up to 64 channels and predicts faster beyond length 700.

\begin{figure}[tb]
\centering
\includegraphics[width=\textwidth]{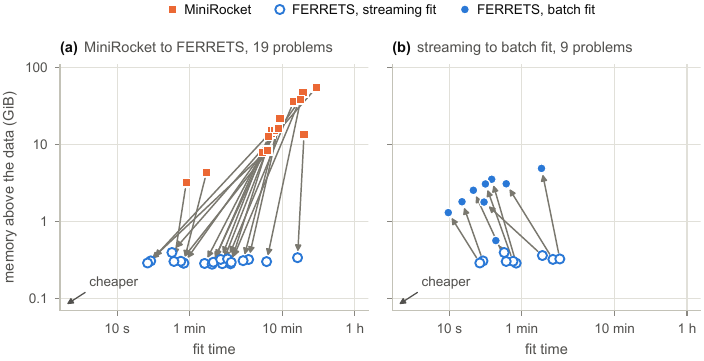}
\caption{MONSTER problems both complete, one arrow each:
(a)~aeon's MiniRocket (ridge head) to streaming FERRETS (open circle),
(b)~streaming to batch (filled circle).}
\label{fig:monster-cost}
\end{figure}

Figure~\ref{fig:monster-cost}a draws one arrow per MONSTER problem, from
MiniRocket to the streaming fit, and every arrow points to a shorter fit and a
lower memory peak. Over
the 19 problems both complete, MiniRocket takes
3.7 times as long in total, and the median peak memory of
FERRETS above the data is 49 times lower. Given that
MiniRocket's ridge head holds its feature matrix in memory, these ratios compare
pipelines. The batch fit is still the one to use where the feature matrix fits in
memory. On the nine MONSTER problems where both run, it takes about half the
streaming fit's time at 7.8 times its peak memory
(Figure~\ref{fig:monster-cost}b), for an accuracy that is not significantly
different. The streaming fit pays elsewhere. Its cache is predicted at
12 times the data at the median, MiniRocket's models are
smaller, and FERRETS takes 1.23 times as long to predict in total
(\href{https://github.com/baraline/ferrets_tsc/blob/main/results/tables/cost-monster.csv}{\texttt{cost-monster.csv}}).

This machine bounds both methods. FERRETS does not complete six of
the 29 MONSTER problems, mostly because of the protocol's guards, not a failed
fit. The disk rule alone excludes LenDB and TimeSen2Crop, and the data of the
full S2Agri problems already exceed the memory budget. Only on the S2Agri-10pc
problems did a fit, without the level word, reach the hour. The gate keeps
MiniRocket off ten (Appendix~\ref{app:protocol}). The
largest problem FERRETS completes holds 1.17 million training series.

Every component that adds accuracy also adds time (\href{https://github.com/baraline/ferrets_tsc/blob/main/results/tables/cost-choices.csv}{\texttt{cost-choices.csv}}).
Without the level word, a model fits in 0.70 to
0.85 of the time, and on Multiverse the cross-channel words cost
about a fifth of the fit. The ridge head is the cheapest head, the streaming head
taking 3.4 times as long to fit on UCR.

\FloatBarrier

\subsection{Accuracy}
\label{sec:accuracy}

\begin{figure}[tb]
\centering
\includegraphics[width=\textwidth]{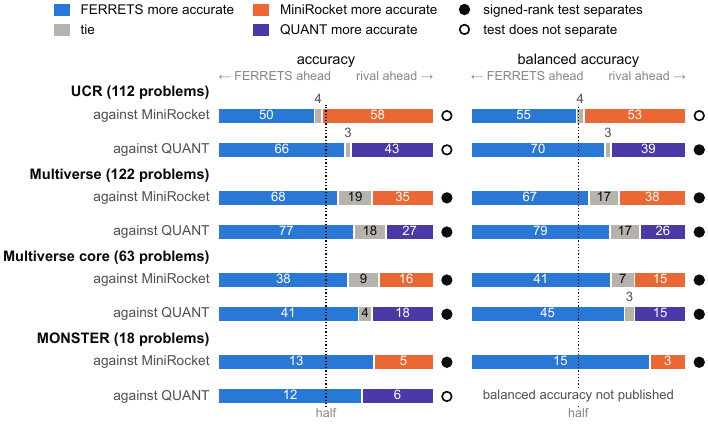}
\caption{Problems won by FERRETS or by the rival, with ties, among those all
three complete, in accuracy (left) and balanced accuracy (right). Filled markers
show where the signed-rank test separates them.}
\label{fig:overall-wins}
\end{figure}

FERRETS is at least as accurate as MiniRocket and QUANT on every archive, and
ahead of both on multivariate data (Figure~\ref{fig:overall-wins}). It is not
significantly different from either on UCR and significantly ahead of both on
Multiverse, as a whole and on its core. At MONSTER, it is significantly ahead of
MiniRocket and not significantly different from QUANT. Outside the development
third, the accuracy verdicts hold on UCR and at MONSTER, but on Multiverse it is
not significantly different from MiniRocket in either measure. Balanced accuracy
changes one verdict, putting FERRETS significantly ahead of
QUANT on UCR, a lead that is not significant outside the third. The
class-balancing rules, which neither competitor applies, explain part of this
lead, not the accuracy leads (Section~\ref{sec:ablation}).

\begin{figure}[tbp]
\centering
\includegraphics{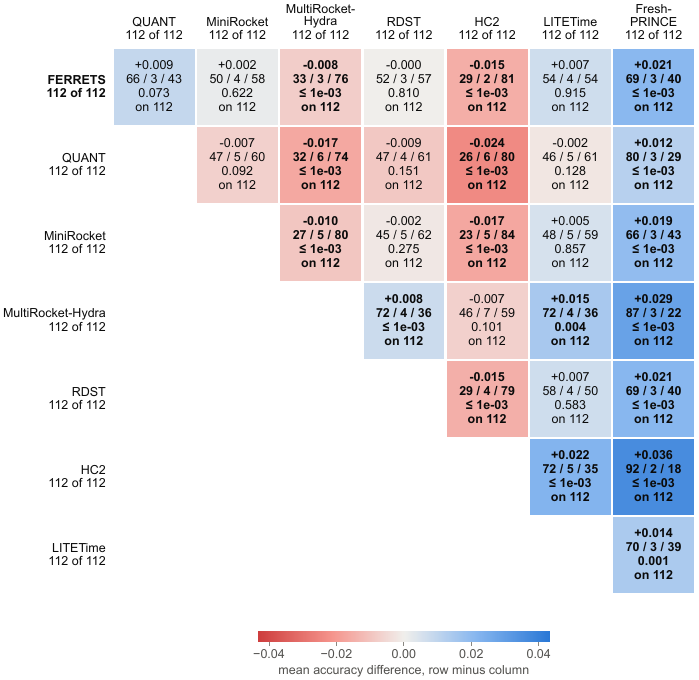}
\caption{Each cell gives the row's mean accuracy minus the column's on the UCR
problems both complete, with wins / ties / losses, Wilcoxon $p$ and problem count,
bold where Holm over all pairs separates the two. A name's number counts the
problems it completes.}
\label{fig:mcm-ucr}
\end{figure}

\begin{figure}[tbp]
\centering
\includegraphics{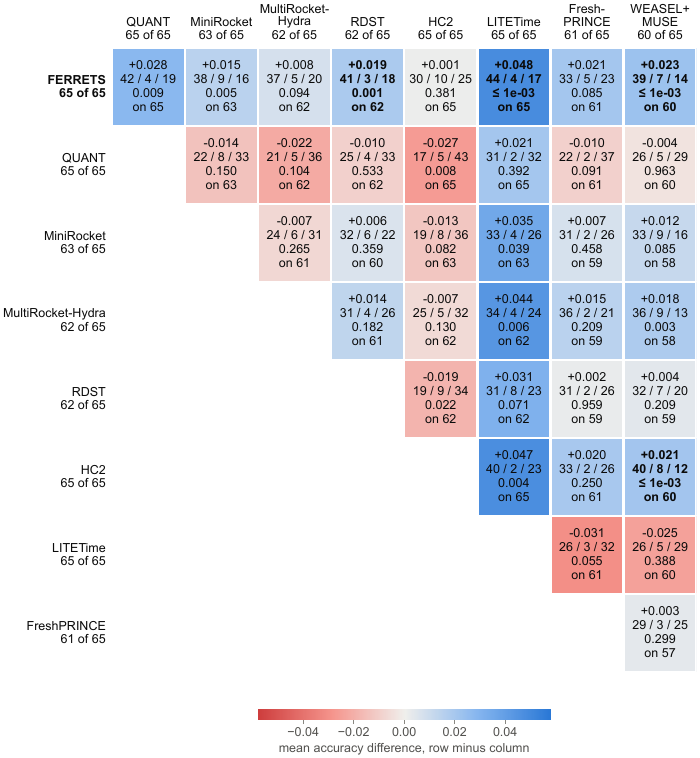}
\caption{Multiverse core, drawn as Figure~\ref{fig:mcm-ucr}, with the
dictionary method WEASEL+MUSE added.}
\label{fig:mcm-mv-core}
\end{figure}

\begin{figure}[tbp]
\centering
\includegraphics{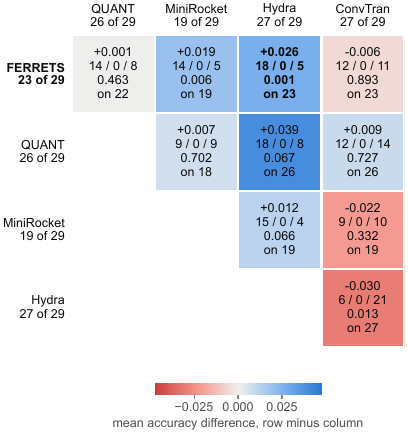}
\caption{MONSTER, drawn as Figure~\ref{fig:mcm-ucr}, with Hydra and ConvTran
standing in for MultiRocket-Hydra and LITETime, which the results workbook lacks.}
\label{fig:mcm-monster}
\end{figure}

Against the strongest method of each family, the matrices of
Figures~\ref{fig:mcm-ucr} to~\ref{fig:mcm-monster} find only HC2 and
MultiRocket-Hydra significantly ahead of FERRETS, and only on UCR. The lead of
MultiRocket-Hydra holds outside the development third, at
1.4 points. Within its own family, no dictionary method
is ahead. FERRETS is significantly ahead of MrSQM, TDE, BORF and cBOSS on UCR and
of WEASEL+MUSE on the Multiverse core, and not significantly different from its
nearest relative WEASEL~2 (\href{https://github.com/baraline/ferrets_tsc/blob/main/results/tables/family.csv}{\texttt{family.csv}}).

\begin{figure}[tb]
\centering
\includegraphics[width=\textwidth]{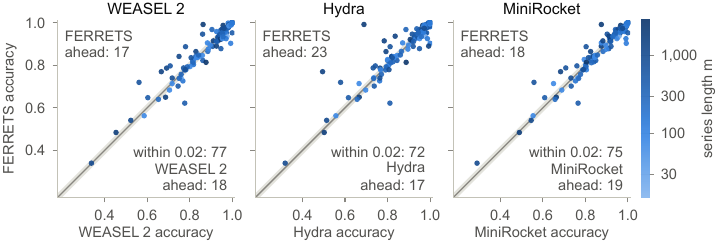}
\caption{Mean accuracy over 30 resamples of FERRETS against
WEASEL~2, Hydra and MiniRocket on the 112 UCR datasets, colored by series
length. The two-point band is a visual convention (1.2 times the
median, over datasets, of FERRETS' standard deviation across resamples).}
\label{fig:pairwise}
\end{figure}

FERRETS' advantage grows with the series length, a significant trend against
MiniRocket and WEASEL~2 (\href{https://github.com/baraline/ferrets_tsc/blob/main/results/tables/ucr-strata.csv}{\texttt{ucr-strata.csv}}). Beyond 700 values, where
MiniRocket starts to predict faster, FERRETS leads it by 2.4
points. On short series with little training data, FERRETS is weakest and trails
MiniRocket by 1.7 points, not significantly, in exploratory
strata. Methods that no test separates are therefore not interchangeable, and in
Figure~\ref{fig:pairwise} their larger gaps run both ways, with the series length
setting the direction.

The file \href{https://github.com/baraline/ferrets_tsc/blob/main/results/tables/mv-channels.csv}{\texttt{mv-channels.csv}} splits the Multiverse problems at eight channels. The
lead over MiniRocket is larger below, though not significantly so, and past eight
channels, where the channel subsets bound the cost, FERRETS is not significantly
different from MiniRocket. On Multiverse as a whole, its wins are many and small
and its losses few and wide. The signed-rank test separates it from MiniRocket,
but the bootstrap interval of the mean difference crosses zero
(\href{https://github.com/baraline/ferrets_tsc/blob/main/results/tables/overall.csv}{\texttt{overall.csv}}).

At MONSTER, FERRETS has the highest accuracy before rounding on
6 of the 13 multivariate problems it
completes. Against each of the seven workbook methods, it is not significantly
different from ConvTran, H-InceptionTime and QUANT and significantly ahead of the
other four (\href{https://github.com/baraline/ferrets_tsc/blob/main/results/tables/monster-pairs.csv}{\texttt{monster-pairs.csv}}). On the problems of at least
100{,}000 series, though, it trails QUANT by 5.4 points over
those both complete.

\begin{figure}[tb]
\centering
\includegraphics[width=\textwidth]{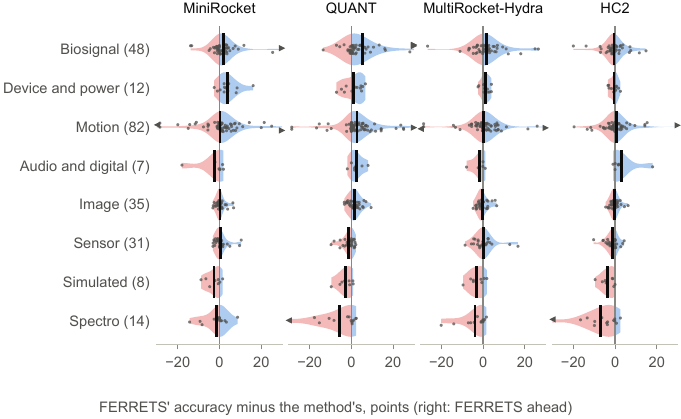}
\caption{FERRETS' accuracy minus each method's by application domain, on the UCR and
Multiverse problems both complete (FERRETS' count in parentheses). Dots are
problems, ticks means and triangles differences beyond the axis.}
\label{fig:domains}
\end{figure}

Figure~\ref{fig:domains} splits the comparison by application domain. FERRETS
does best on biosignals and on device and power data, the two UCR domains with
the longest series, and worst on simulated data, the shortest, and on spectra.
The groups are small, so this reading is descriptive. Corrected over the figure's
cells, only the leads over QUANT on biosignal and motion data are significant
(\href{https://github.com/baraline/ferrets_tsc/blob/main/results/tables/domains.csv}{\texttt{domains.csv}}). HC2 stays ahead on biosignals in both archives, and
QUANT and HC2 on sensor data.

MONSTER's seven workbook methods enter for their accuracy only, measured on its
authors' budgets. FERRETS' batch fit has no
result on eight of the 133 Multiverse problems.

\FloatBarrier

\subsection{Foundation models}
\label{sec:tsfm}

\begin{figure}[!htb]
\centering
\includegraphics{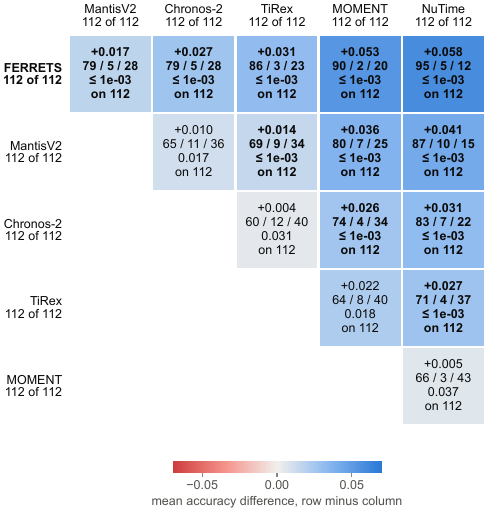}
\caption{FERRETS and the five foundation models on UCR's published split,
112 datasets, drawn as Figure~\ref{fig:mcm-ucr}.}
\label{fig:mcm-tsfm}
\end{figure}

Figure~\ref{fig:mcm-tsfm} compares FERRETS on UCR's published split with the
latest versions of the five time-series foundation models that have published
results on the whole archive, MantisV2~\cite{feofanov2026mantisv2}, Chronos-2~\cite{ansari2025chronos2},
TiRex~\cite{auer2025tirex}, MOMENT~\cite{goswami2024moment} and
NuTime~\cite{lin2024nutime}. Used as frozen encoders, their pretrained weights left
unchanged, under a logistic-regression head, none of them reaches FERRETS, which
is significantly more accurate than each one, corrected over the matrix's pairs.
Even the most accurate, MantisV2, trails it by 1.7 points and is less
accurate on average than MiniRocket and QUANT.

The evidence is one split, the only one the models' results cover, taken
from~\cite{feofanov2026mantisv2} with the report's preprocessing and head, and
37 of its 112 datasets belong to the development third
(Appendix~\ref{app:protocol}).

\FloatBarrier

\subsection{Ablation}
\label{sec:ablation}

\begin{figure}[!tb]
\centering
\includegraphics[width=\textwidth]{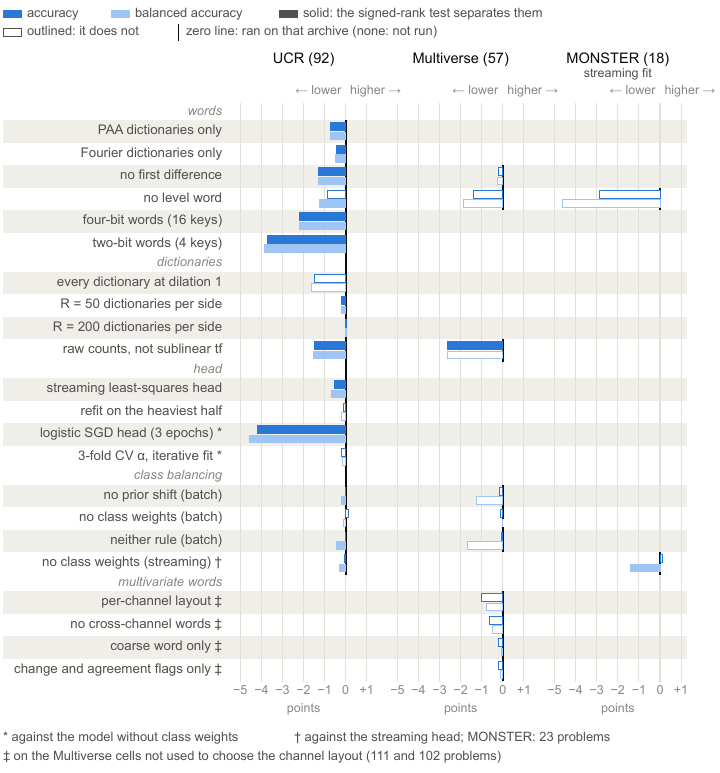}
\caption{Points of accuracy (dark) and balanced accuracy (light) each variant gains or loses against the model it changes, solid where the signed-rank test separates them.}
\label{fig:ablation-card}
\end{figure}

Figure~\ref{fig:ablation-card} and the files \href{https://github.com/baraline/ferrets_tsc/blob/main/results/tables/ablation-ucr.csv}{\texttt{ablation-ucr.csv}}
and \href{https://github.com/baraline/ferrets_tsc/blob/main/results/tables/ablation-mv-monster.csv}{\texttt{ablation-mv-monster.csv}} pair each variant with the shipped model on the
ablation's problems, which contain the development third the defaults were chosen
on (Appendices~\ref{app:protocol} and~\ref{app:ablation}). No variant is significantly
more accurate, and removing most parts lowers accuracy or balanced accuracy. Where
the test cannot separate a drop from zero, the choice rests on cost or design.

Accuracy rests on the dictionary, on the ridge head and, for long series, on the
level word. On UCR, keeping one word
source only, dropping the first difference, shortening the words, using raw
counts or keeping fifty dictionaries per side each lowers accuracy and balanced
accuracy significantly, while two hundred dictionaries gain nothing. Dilation 1
alone loses accuracy, not significantly, and the dilations stay by design, for
words spread over a window. A logistic
head loses 4.2 points of accuracy against the unweighted ridge
head, and the streaming head also loses accuracy, the price of streaming. We keep
the level word despite its time cost. Removing it lowers
accuracy and balanced accuracy on every archive, significantly only for UCR's
balanced accuracy, and the loss lies mostly on series longer than 700 points,
where FERRETS leads.

The class-balancing rules buy balanced accuracy, not accuracy. On UCR, removing
the shift, both batch rules or the streaming weights lowers balanced accuracy
significantly, by 0.4 points for both batch rules, and removing
the streaming weights does the same at MONSTER. On Multiverse, the loss is not
significant. Adding the batch weights to the shift alone is not significantly
different on either measure, so the weights stay as a design choice.

The multivariate choices are the least supported part of the design. Keying each
channel alone, dropping the cross-channel words or keeping only part of them each
lowers accuracy, not significantly, and the problems that gain do not recur
across resamples more often than chance (Appendix~\ref{app:ablation}). The
channel layout stays for its cost bound, the cross-channel words for a small gain
that costs fit time, and the flags because their keys decode
(Figure~\ref{fig:mv-words}), not for a measured gain.

We leave the optional bag transforms off, since they help nowhere. On UCR,
inverse document frequency, numerosity reduction and all three together are
significantly worse, and no addition is significantly different on Multiverse or
at MONSTER. Even the best combination per dataset, chosen on held-aside
resamples, gains 0.07 points, not significantly different from the
defaults.

\FloatBarrier

\section{Conclusion}
\label{sec:conclusion}

FERRETS is a label-free symbolic transform with a bounded output and a cost
linear in series length (Section~\ref{sec:transform}). A linear head on its
counts fits in time linear in the training set above 3{,}000
series and predicts at constant cost per series. Under a ten-thread budget,
FERRETS is the fastest of the eleven methods timed on UCR (Section~\ref{sec:cost}).
On the MONSTER problems both complete, MiniRocket takes 3.7
times as long to fit as the streaming FERRETS classifier, which is significantly
more accurate (Sections~\ref{sec:cost} and~\ref{sec:accuracy}).

FERRETS' accuracy is
not significantly different from MiniRocket's and QUANT's on UCR
and significantly ahead of theirs on Multiverse, by
0.9 and 2.6 points, though not of MiniRocket's
outside the development third.

\paragraph{Limitations.}
Related problems fall on both sides of the development third, selection optimism
is not estimated, and the tests treat them as independent
(Appendix~\ref{app:protocol}). We use one machine and, outside UCR, one seed. The transform was tested only under its own
heads, so cost ratios and accuracy leads compare whole pipelines. Neither
MiniRocket nor QUANT applies the class-balancing rules, which explain part of the
balanced-accuracy leads.

The batch fit stops where its feature matrix outgrows memory. The predicted
streaming cache is 12 times the data on disk at the median,
and the largest MONSTER problems stay out (Section~\ref{sec:cost}). Start-up and
single-series latency are not measured (Appendix~\ref{app:cost}). MiniRocket fits faster on smaller Multiverse problems and
predicts faster beyond 700 values (Section~\ref{sec:cost}). The Multiverse lead
is larger with at most eight channels (Section~\ref{sec:accuracy}). On shorter
series with few training series, FERRETS trails MiniRocket by
1.7 points, not significantly, and on the completed MONSTER
problems of at least 100{,}000 series, it trails QUANT by 5.4
points (Section~\ref{sec:accuracy}).

\paragraph{Reproducibility.}
The package \texttt{ferrets} imports aeon, numpy, numba, scipy and scikit-learn, and
its transformer's \texttt{transform} method returns the weighted rows
(Definition~\ref{def:tfidf}). The
repository%
\footnote{\url{https://github.com/baraline/ferrets_tsc}}
holds the package with the result rows behind the figures and tables, and the
scripts that produced those rows.

\appendix
\def\theHsection{appendix.\Alph{section}}%
\section{Protocol and cost details}
\label{app:protocol}

This appendix completes Sections~\ref{sec:protocol} and~\ref{sec:cost}.

\subsection{Runs}

We measured all times on one desktop with an AMD Ryzen 5 5500 (six cores,
12 hardware threads), 64\,GiB of memory and an NVMe SSD under Windows. The
software was Python~3.14.6, aeon~1.5.0~\cite{aeon},
scikit-learn~1.8.0~\cite{sklearn} and numba~\cite{numba}. MrSQM has no Windows
build, so it ran on the same machine under WSL2, the Linux environment built
into Windows, with the older scikit-learn~1.6.1 it requires. Every method gets the
same thread budget, set through the libraries' thread caps and the estimator's
own option. MrSQM gets it too but has no parallel implementation, so its core
uses one thread.

\paragraph{Outcomes.}
A cell, one dataset and one method, runs in its own subprocess over all its
folds. It completes, or it fails when a fold overruns the hour or raises an
error. After an overrun, the later folds are not run. Only these measured
failures count against a method. A cell can also be skipped on the gate's
prediction, excluded by the disk rule, or not attempted. A cell is not attempted
when the method already failed on the dataset or did not complete a smaller size
on the same axis of the size grid. A streaming fit is also not attempted when
the same fit without the level word timed out.

\paragraph{Feasibility gate.}
The gate predicts a cell's peak memory, fit time and prediction time. Its
models are log-log fits, one per method, of these costs against the number of
series and the number of input values, fitted on the cells this machine
completed. The gate covers every MONSTER cell, the size grid and the real
problems timed with it, the Multiverse copies of MONSTER problems and the
slowest methods (WEASEL~2, MrSQM, cBOSS, TDE and WEASEL+MUSE). It skips a cell
if the method already failed on that dataset, if a fold's predicted time reaches
the hour, or if the predicted peak memory reaches 56\,GB, about 81\,\% of the
machine's memory. At MONSTER it skips MiniRocket on ten
problems. Each of them has a predicted peak of 3 times the
memory budget or more, except DREAMERA and DREAMERV at 1.7
times.

\paragraph{Disk rule.}
A cell is excluded when its data or its streaming cache would not fit the free
disk space. The space counted is the larger of two values, the space the disk
had free when the cell was decided and 60\,GB, which we take as the free space
of a healthy disk. The cache may use at most 50\,\% of that space, less the
cell's download. TimeSen2Crop and LenDB fit the memory budget, but their caches,
about 368 and 244\,GB, fit neither the allowance on the
128\,GB the disk had free when they were decided nor that of a healthy
disk. The gate predicts LenDB well within the hour and TimeSen2Crop close to it,
so the disk rule alone excludes these two problems. The full S2Agri problems are
skipped on predicted memory, since their data alone take 114\,GB as
double-precision floats, above the budget. The S2Agri-10pc problems are not
attempted because the fit without the level word reached the hour.

\paragraph{Published results.}
The published results are those of the archives (Table~\ref{tab:archives}). For
Multiverse they are the tables of \texttt{github.com/aeon-toolkit/multiverse} at
commit \texttt{5fdf986}. Those runs had the archives' own
budgets~\cite{multiverse,MONSTER}, so the problems they complete differ from
ours. They carry no time or memory measured on this machine, so we re-ran the
UCR methods that have a published column for time and memory only. Of the
Multiverse problems, 18 are also MONSTER problems, and Multiverse's split of
each is MONSTER's fold~0. The MONSTER results come from the workbook
\texttt{monster\_final\_results.xlsx} of the MONSTER code
repository~\cite{monsterresults}. We matched its problems to the hub's by name,
or by labels and test indices, except the full S2Agri problems. The workbook's
methods keep their authors' preprocessing, which we cannot state, while FERRETS
and MiniRocket read the hub's raw arrays.

\paragraph{Settings of the methods we run.}
MiniRocket and Hydra run as aeon ships them, at their default sizes, with a
ridge classifier whose penalty is cross-validated on the feature matrix held in
memory. MiniRocket's default of 10{,}000 kernels is a capacity setting, which we
did not tune. The costs of both methods are thus those of this head, not of the
logistic head their authors recommend for large training
sets~\cite{MiniRocket,hydra}. BORF is aeon's transform at its defaults under the
same ridge classifier. Its counts are cast to float, since
scikit-learn~1.8.0's ridge fitted on integer features predicts a single class.
WEASEL+MUSE~\cite{muse} runs with its paper's logistic head. With aeon's default
ridge head it scores near the chance rate on nearly every equal-length problem
of a check run here (\texttt{results/uea\_muse.csv}). The batch fit of FERRETS
also runs on the nine MONSTER problems whose feature matrix fits in 8\,GB, as
estimated beforehand. The MONSTER accuracies are those of the streaming fit.

\paragraph{Dependence and seeds.}
Multiverse holds groups of problems from one source (for example UIPRMD,
IRDS, KERAAL and KIMORE), which the tests treat as independent. Averaging each group
before the primary tests leaves 88 groups for the
122 problems that FERRETS, MiniRocket and QUANT complete
(62 groups for 63 core problems), and no
verdict changes. Related UCR datasets that share series, such as FaceAll and
FacesUCR, fall on both sides of the development third. We did not estimate the
selection optimism, the gain that a choice made on the third can carry over to
related problems outside it. We fitted the ablation's 92 UCR datasets again at
resamples 1 to 9 with five seeds. The mean accuracy lies between 0.8728
and 0.8736, and every seed gives the primary verdicts of seed~0. We did
not fit the Multiverse and MONSTER problems again with other seeds.

\paragraph{Design data.}
We chose FERRETS' defaults on a \emph{development third} of each archive, at
most a third of its problems whatever the split: 37 of the 112 UCR
datasets, 44 of the 133 Multiverse problems and
9 of the 29 MONSTER problems. We drew these problems at
random, with seed~0, among those on which every variant of the ablation has
results. The draw is stratified by series length, in three bands ($m \le
150$, $150 < m \le 700$ and $m > 700$), each in proportion to its size. MONSTER
was drawn first. Multiverse splits a problem that MONSTER also carries as
MONSTER's fold~0, so the five such problems of the MONSTER third count
in Multiverse as well, which leaves 39 Multiverse problems to draw.
The choices read the UCR third at resamples 1 to 9, the Multiverse third at the
published split and the MONSTER third at fold~0. For the channel layout they
also read resamples 1 and 2 of the Multiverse problems with $C \ge 9$. A variant
replaces a default only if it is significantly better than the shipped model in
accuracy or balanced accuracy, not significantly worse in the other, and better
by at least 0.02. We applied this rule with the tests and Holm families of
\href{https://github.com/baraline/ferrets_tsc/blob/main/results/tables/ablation-ucr.csv}{\texttt{ablation-ucr.csv}} and \href{https://github.com/baraline/ferrets_tsc/blob/main/results/tables/ablation-mv-monster.csv}{\texttt{ablation-mv-monster.csv}}, restricted to the
third. None of the 43 variants is significantly better than
the shipped model in accuracy or balanced accuracy, so every default stays. The
streaming head's step and number of epochs were chosen on the UCR development
third alone. The ablation's problems include those of the third, which the
per-dataset files of Appendix~\ref{app:results} flag.
The memory and one-thread runs of Section~\ref{sec:cost} use 20 other UCR
datasets, listed in \texttt{results/ucr\_memory\_runs.csv}, and the ablation's 92
UCR datasets are the rest.

\paragraph{Outside the development third.}
We repeat the primary comparison on the problems outside the third that
FERRETS, MiniRocket and QUANT complete: 75 UCR datasets at
every resample, 80 Multiverse problems on the published split
and 9 MONSTER problems at every fold. The tests run
with Holm's correction over the two rivals in each archive and measure. On UCR,
FERRETS is not significantly different from MiniRocket or QUANT in accuracy or
balanced accuracy. On Multiverse, it is significantly ahead of QUANT in both
measures and not significantly different from MiniRocket. At MONSTER, it is
significantly ahead of MiniRocket in accuracy and balanced accuracy and not
significantly different from QUANT in accuracy. The full archives give the same
verdicts (Section~\ref{sec:accuracy}), except that on them FERRETS is also
significantly ahead of MiniRocket on Multiverse and of QUANT in balanced
accuracy on UCR.

\subsection{Cost measurement}
\label{app:cost}

The values behind Section~\ref{sec:cost} are in the files \href{https://github.com/baraline/ferrets_tsc/blob/main/results/tables/cost-ucr.csv}{\texttt{cost-ucr.csv}}, \href{https://github.com/baraline/ferrets_tsc/blob/main/results/tables/cost-mv.csv}{\texttt{cost-mv.csv}}, \href{https://github.com/baraline/ferrets_tsc/blob/main/results/tables/cost-monster.csv}{\texttt{cost-monster.csv}}, \href{https://github.com/baraline/ferrets_tsc/blob/main/results/tables/cost-scaling.csv}{\texttt{cost-scaling.csv}} and \href{https://github.com/baraline/ferrets_tsc/blob/main/results/tables/cost-choices.csv}{\texttt{cost-choices.csv}} (Appendix~\ref{app:results}).

\paragraph{Units and timing.}
Timings measure throughput on warm code, after a fit and a prediction on a small problem have absorbed the compilation. We did not measure start-up or the latency of a single series. Memory above the data is the main process's peak resident size minus its size once the data were loaded, and total memory is the peak itself. Neither counts worker processes, so the memory of WEASEL+MUSE, whose aeon implementation starts joblib workers, is understated. The benchmark samples the resident size every 20\,ms during each run. The memory runs behind \href{https://github.com/baraline/ferrets_tsc/blob/main/results/tables/cost-ucr.csv}{\texttt{cost-ucr.csv}} instead read the peak from the operating system's counter, so the memory columns of that file do not compare with those of \href{https://github.com/baraline/ferrets_tsc/blob/main/results/tables/cost-mv.csv}{\texttt{cost-mv.csv}} and \href{https://github.com/baraline/ferrets_tsc/blob/main/results/tables/cost-monster.csv}{\texttt{cost-monster.csv}}. Model size is the size of the pickled estimator. An MB is a million bytes, and a GiB is 1{,}073.74\,MB. A MONSTER cost is the mean over the five folds.

\paragraph{UCR.}
Every UCR cost is at resample~10. Hydra, cBOSS and 85 of the 89 TDE cells were timed in one long-running process without a recorded thread count. MiniRocket's result file has no thread column. On UCR, TDE and cBOSS ran without the fit-time limit that they get on the size grid. The memory and model columns come from the memory runs, on 19 of their 20 datasets, with at most 1{,}000 training series.

\paragraph{The size grid.}
Each cell of the size grid is a synthetic five-class problem. Per class and channel, a series is a template of sinusoids and a Gaussian bump, shifted circularly and noised. The test set has 50\,\% as many series as the training set. Every completed run classifies perfectly, so data-dependent costs may sit near their best case. A cell's time is the median of three timed repeats of the same fit, with the same draw of the dictionaries. TDE and cBOSS run under aeon's contract, a fit-time limit that we set to 50 minutes, below the hour cap. In \href{https://github.com/baraline/ferrets_tsc/blob/main/results/tables/cost-scaling.csv}{\texttt{cost-scaling.csv}}, the last block is the per-channel ablation. The channel slopes span 4 to 256 channels, except that of WEASEL+MUSE, which spans three grid points, the last with a single repeat.

\FloatBarrier

\section{Results files}
\label{app:results}

The folder \texttt{results/per\_dataset/} of the repository holds the accuracy and balanced accuracy behind
Section~\ref{sec:experiments}, one row per problem, in three files: \texttt{ucr.csv} for the 112 UCR datasets,
\texttt{multiverse.csv} for the 133 Multiverse problems and \texttt{monster.csv} for the 29 MONSTER problems.
Their column \texttt{development} flags the problems of the development third
(Appendix~\ref{app:protocol}).
Beside FERRETS, the three files list MiniRocket and QUANT. The UCR and Multiverse files add Hydra and HC2, and the
UCR file also WEASEL~2. The MONSTER file adds the seven methods of the results workbook. The other methods
compared have their per-dataset results in the archives' published files (Table~\ref{tab:archives}), except BORF,
which we ran and whose results are in the repository.
A UCR value is the mean over the 30 resamples of the dataset (resample~0 being the published split), a MONSTER
value the mean over its five folds, and a Multiverse value the result at the published split. We ran FERRETS on
the three archives (batch fit on UCR and Multiverse, streaming fit at MONSTER), MiniRocket and Hydra on
Multiverse, and MiniRocket at MONSTER. Every other column is a published result, at MONSTER the
workbook's~\cite{monsterresults}, which gives balanced accuracy for ConvTran, FCN, H-InceptionTime and TempCNN
only. The files read the workbook's first sheet, the only one that holds all seven methods. An empty cell is a missing published result or a cell run here without a result. For a cell run here, the
method's note column gives its state under the rules of Appendix~\ref{app:protocol} (predicted skip, not
attempted, disk exclusion) or its failure (timeout, out of memory, near-constant channels refused by aeon's input
check, series shorter than the method accepts, unequal lengths).

The tables behind Section~\ref{sec:experiments} are the files of \texttt{results/tables/}, one per table. Their
README gives their columns, conventions and Holm families. Among them, \href{https://github.com/baraline/ferrets_tsc/blob/main/results/tables/mv-core.csv}{\texttt{mv-core.csv}} and \href{https://github.com/baraline/ferrets_tsc/blob/main/results/tables/mv-all.csv}{\texttt{mv-all.csv}}
compare every published method on the Multiverse core and on all 133 problems. Among the methods of these files,
MultiRocket-Hydra is Hydra combined with MultiRocket~\cite{hydra}. The workbook's methods are
ConvTran~\cite{foumani2024convtran}, FCN~\cite{wang2017fcn}, H-InceptionTime~\cite{ismailfawaz2022hccf}, Hydra,
QUANT, TempCNN~\cite{pelletier2019tempcnn} and ET, an ensemble of extremely randomized
trees~\cite{geurts2006extremely}. The Multiverse files add FreshPRINCE~\cite{middlehurst2022freshprince}.

\section{Ablation variants}
\label{app:ablation}

This appendix describes the variants behind Figure~\ref{fig:ablation-card}, whose values are in the files
\href{https://github.com/baraline/ferrets_tsc/blob/main/results/tables/ablation-ucr.csv}{\texttt{ablation-ucr.csv}} and \href{https://github.com/baraline/ferrets_tsc/blob/main/results/tables/ablation-mv-monster.csv}{\texttt{ablation-mv-monster.csv}}. The shipped model is FERRETS at its default settings.
Holm's correction runs over the 46 tests of the first file and over the 40 tests of the second, which pools its
three blocks in one family.

\paragraph{The cells.} The ablation runs on 92 UCR datasets, all but the 20 of the memory runs, at resamples 1 to 9.
On Multiverse it runs on the 57 problems of the core that are not MONSTER problems, at the published split. At
MONSTER it runs on 18 problems at fold~0, and its streaming class-weight row on the 23
problems that both streaming fits complete. The channel-layout rows read resamples 1 and 2 of
73 of the 76 Multiverse problems with $C \ge 9$. The per-channel row adds the published split
of 38 other problems outside MONSTER, 111 in all, and the cross-channel rows that of 29 of
the 57, 102 in all.

\paragraph{The variants.} Four- and two-bit words shorten the PAA and Fourier words only. The pruning
row drops the columns that occur in fewer than two training series. The logistic head is three epochs of averaged
stochastic gradient descent on the log loss, with the streaming head's $\ell_2$ penalty per series, so it is not a
converged fit. The cross-validated head picks its penalty by three-fold cross-validation on the iterative route at
every size. Neither head takes class weights, so both pair with the batch fit without them. The streaming
class-weight row pairs with the shipped streaming fit. The oracle picks, for each dataset, the best combination of
the bag transforms on the same resamples on which it is scored. The last row of \href{https://github.com/baraline/ferrets_tsc/blob/main/results/tables/ablation-ucr.csv}{\texttt{ablation-ucr.csv}} instead
chooses the combination on resamples 1 to 4 and scores it on 5 to 9. The refit scores each family of each side by
the $\ell_2$ norm of its head weights over all classes, keeps the heavier half and fits again. All but three of the
92 UCR datasets have fewer series than the 3{,}000 at which the batch head switches routes, so the other batch rows
compare fits of the exact ridge on all but those three.

\paragraph{Gains problem by problem.} Over the 72 problems with $C \ge 9$ that have both
resamples 1 and 2, the cross-channel words gain at least 2 points at both resamples on only
two problems, where chance alone would give 1.1.
The level word's gains recur on 12 problems against 5.5 by
chance, a significant excess under a one-sided hypergeometric test on the count. This test can thus detect a
per-problem effect.

\section{Derivations}
\label{app:derivations}

Sections~\ref{sec:contributions} and~\ref{sec:classification} rest on the proofs and long derivations gathered here.

\subsection{Flat layout}
\label{app:column-example}

\paragraph{Columns.} The blocks are laid end to end: side after side, then family
after family (Fourier before PAA, by increasing window), then dilation after
dilation, $2^{8}$ columns each. The block of the level words of a (side, window,
dilation) follows the word block of the first dictionary with that window and
dilation. For a dictionary $\delta$, let $o_\delta$ be the first column of its
family and $q_\delta$ the rank of its block in that family. This rank is the rank
of its dilation, counted from 0, plus the number of level blocks laid before it.
The column of a word $u$ of key $\kappa(u)$ is then
\begin{equation*}
\mathrm{col}(\delta, u) = o_\delta + 2^{8}\,q_\delta + \kappa(u).
\end{equation*}
For $C \ge 2$, the families of one source and window are also ordered by their
channels. Each dilation of a family holds $g + 3$ word blocks, $g$ being the
number of channels the family reads. These blocks hold the channel-wise words in
channel order, then the coarse word, the change flags and the agreement flags.
The level block of the family's (side, channels, window, dilation) follows them
if the family is the first to key it. The rank $q_\delta$ then counts all these
blocks.

\paragraph{Decoding.} Decoding inverts the arithmetic. The list of kept columns
maps a head feature to its column. A binary search over the family offsets gives
the family and its offset $o$. The block is then
$q = \lfloor (\mathrm{col} - o) / 2^{8} \rfloor$, and the eight bits are
$\kappa = (\mathrm{col} - o) \bmod 2^{8}$.

\subsection{Bits, ties and rounding}
\label{app:ties}

This subsection supports the paragraphs ``Draws'' and ``From values to bits'' of
Section~\ref{sec:transform}.

The keying kernel decides every bit by a comparison in double precision. It takes
a window's mean and variance from prefix sums of its residue chain, centered on
the chain's mean. A residue chain holds the values of the series that lie one
dilation apart (Appendix~\ref{app:keys-proof}). When the variance so obtained is
at most $\rho_{1}$ times the chain's variance, or $\rho_{2}$ times the window's
mean square about the chain mean, the kernel recomputes both moments in two
passes over the window. A PAA bit compares a segment sum with an eighth of the
sum of the window's eight segment sums. A Fourier bit compares the window's
coefficient, scaled by $1/(w\sigma)$, with its cut. A level bin counts the cuts
strictly below the moment. The PAA and Fourier words of a flat window
(Definition~\ref{def:word}) take fixed keys.

Every comparison with a cut is strict, so a value equal to its cut gives a $0$, or
the lower bin. Ties arise on quantized series. The rounding of the prefix sums
depends on where a window lies in its chain and grows with the chain's length, so
rounding alone could give identical windows different keys. FERRETS decides ties
by the window's values instead. The PAA threshold is computed from the window's
own segment sums, each added in a fixed order. Each stored cut is raised above the
values it was drawn from by a margin. A Fourier cut is raised by
$\epsilon_{\mathrm{F}}$, a standard-deviation cut $\theta^{\sigma}_j$ to
$\theta^{\sigma}_j(1+\epsilon_{\sigma})$, and a mean cut $\theta^{\mu}_j$ to
\begin{equation*}
\theta^{\mu}_j + \epsilon_{\mu}\big(|\theta^{\mu}_j| + \theta^{\mu}_{\max} - \theta^{\mu}_{\min}\big),
\end{equation*}
with $\theta^{\mu}_{\max}$ and $\theta^{\mu}_{\min}$ the largest and smallest cut of its own set. A set whose
mean cuts are all $0$ keeps them unmoved, so a window of mean exactly $0$ takes
the first or the last bin, as rounding decides. Every standard-deviation cut at most
$\epsilon_{0}$ is set to $0$, the standard deviation that the level word gives a
flat window. The constants are
\begin{equation*}
\rho_{1} = 10^{-6}, \quad \rho_{2} = 10^{-8}, \quad \epsilon_{\mathrm{F}} = 10^{-6},
\quad \epsilon_{\mu} = 10^{-10}, \quad \epsilon_{\sigma} = 10^{-5},
\quad \epsilon_{0} = 10^{-8}.
\end{equation*}
We set them once and did not tune them. The constant $\epsilon_{0}$ is also the
flat threshold of Definition~\ref{def:word}.

Ties are common on quantized series (\href{https://github.com/baraline/ferrets_tsc/blob/main/results/tables/ties.csv}{\texttt{ties.csv}}), and
3.7\,\% of the training windows of the word blocks have an
identical twin. We checked that no group of identical windows of the word or
level blocks receives more than one key. The check covers the training and test
parts of the published split of the 112 UCR datasets, synthetic series of integer
levels at two lengths and two MONSTER problems of integer counts. It covers the
univariate kernel alone, on series no longer than UCR's longest, while the
rounding that the margins must exceed grows with the series length. We did not
test the tie property of the multivariate keying, on which the change and
agreement flags of Section~\ref{sec:contrib-mv} rely.

\subsection{Proof of the cost of the keys}
\label{app:keys-proof}

This subsection proves Proposition~\ref{prop:complexity}.

The windows of a dilated sequence lie on its $d$ residue chains, the sequences
$v = (x_i, x_{i+d}, x_{i+2d}, \ldots)$ for a start $i \le d$, of length
$L \approx m/d$. WEASEL~2 also reorders a series into these
chains~\cite{schafer2023weasel}. The chain lengths sum to $m$, so it suffices to
key one chain in $O(L)$ time. Prefix sums of the chain minus its mean, and of the
squares of that difference, give the mean and variance of every window by two
subtractions~\cite{UCRSuite}. A window that fails the variance test of
Appendix~\ref{app:ties} is recomputed in two passes, in $O(w)$ with $w \le 64$, so
the chain stays in $O(L)$ even when every window is recomputed. Within its chain a
window is $w$ consecutive values, so the dilation adds no work. On a univariate
series, the dictionaries of a side that share a dilation read the same chains.
The prefix sums are thus built once per series and dilation, and the moments once
per window length.

A PAA word needs eight segment sums. Let $A_t$ be the sum of $w/8$ consecutive
chain values, which takes at most seven additions per position. The segment sums
of the window at $t$ are then $A_t, A_{t+w/8}, \ldots, A_{t+7w/8}$. The window's
threshold is their total times $1/8$, and its word costs eight comparisons.

A Fourier word needs coefficients $1$ to $4$. Coefficient $h$ of the z-normalized
window $Z$ of Definition~\ref{def:word} is
$\hat z_h = \frac{1}{w}\sum_{j=0}^{w-1} z_j\,e^{-2\pi\mathrm{i}\,hj/w}$. Since
$\sum_{j=0}^{w-1} e^{-2\pi\mathrm{i}\,hj/w} = 0$ for $1 \le h \le 4 < w$, the mean
drops out, and for the window at $t$,
$\hat z_h = \frac{1}{w\sigma}\sum_{j=0}^{w-1} v_{t+j}\,e^{-2\pi\mathrm{i}\,hj/w}$.
With $M_h(t) = \sum_{a<t} v_a\,e^{-2\pi\mathrm{i}\,h a/w}$, that sum is
$e^{2\pi\mathrm{i}\,h t/w}\,(M_h(t+w) - M_h(t))$. Four modulated prefix sums, one
rotation per coefficient and one division by $w\sigma$ thus serve all windows, and
a key costs eight comparisons. The scheme is the sliding Fourier transform that
BOSS~\cite{BOSS} applies to SFA words, without its step-by-step update. The
cosines and sines of each coefficient form a table built once per window length.

A level word reads the mean and variance already computed and costs two binary
searches among fifteen sorted cuts, four comparisons each. One pass over the keys
fills $2^{8}$ counters. A mask of $2^{8}$ bits marks the keys seen, and the
counters are read out in key order through it, with no sort and no hash table.
The read-out clears the counters and the mask for the next block. Keying $n$
series under the $V$ dictionaries and their level blocks repeats these steps for
each series, dictionary and chain, which gives the time $O(n\,V\,m)$. The number
$V$ is at most the $2R$ draws, so the time is linear in $n\,m$. The number $V$
nears $2R$ as $m$ grows, since a longer series admits more dilations. The time
bound holds without prefix sums, since a window computed directly costs $O(w)$
with $w \le 64$. The prefix sums make the cost of a window's moments and Fourier
coefficients independent of $w$.

With several channels, a channel-wise word that two dictionaries of a side would key
alike is keyed once, as long as the shared words fit a table of 1~MB. The flags
compare the channel-wise words already keyed, and the Fourier coarse word takes
their leading bits.

\subsection{The choice of the shift}
\label{app:lambda}

This subsection supports the paragraph ``Choosing the shift'' of
Section~\ref{sec:contrib-imbalance}.

The shift is scored at the head's selected $\alpha$, on the same out-of-sample
decisions for the training series that chose $\alpha$. These decisions are the
exact leave-one-out decisions of the dual route, or the decisions on the 25\,\%
hold-out of the iterative route. Each $\lambda$ is scored by
$\varsigma(\lambda) = (\text{accuracy} + \text{balanced accuracy} + \text{macro
F1})/3$ of its shifted decisions against the training labels. The chosen
$\lambda$ is the smallest with
\begin{equation*}
\varsigma(\lambda) \ge \varsigma^{*} - \sqrt{\varsigma^{*}(1 - \varsigma^{*})/n'},
\end{equation*}
$\varsigma^{*}$ being the best score and $n'$ the number of decisions scored. The
margin is the standard error that a proportion $\varsigma^{*}$ of $n'$ independent
trials would have, the form Breiman et~al.~\cite{Breiman1984} use for an error rate
in their one-standard-error rule. Since $\varsigma$ is not such a proportion and
the decisions are not independent, the margin is a heuristic, neither the standard
error of $\varsigma$ nor a test. The smallest $\lambda$ takes the role of the
rule's simplest tree. The value $\lambda = 0$ thus stands unless a larger value
beats it by more than the margin, which shrinks as $n'$ grows.

\subsection{The two batch routes in detail}
\label{app:ridge-routes}

This subsection supports the two batch routes of Section~\ref{sec:contrib-scale}.

Both routes fit the weighted ridge of Section~\ref{sec:classification}. The dual
route takes the weighted leave-one-out squared error of each of its five penalties
in closed form, from one eigendecomposition of the weighted Gram matrix of the
centered rows~\cite{rifkin2007notes}. It follows scikit-learn's computation for a
ridge with sample weights and an unpenalized intercept~\cite{sklearn}. It keeps
the weighted rows, their column means and the coefficients in float32, and
computes the Gram matrix and its eigendecomposition in float64. ``Exact'' thus
means the closed-form minimizer for the rows as stored, up to rounding.

In the iterative route, LSQR~\cite{paige1982lsqr} stops a class after 200 steps,
or earlier once one of its first two stopping rules holds for every penalty, both
rules at the tolerance of Table~\ref{tab:head}. Our rule on the residual leaves
out the penalty term that LSQR's own rule includes, so it is met no later than
LSQR's. The route picks the penalty with the most correct decisions on a hold-out
of 25\,\% of the training series, scikit-learn's default share~\cite{sklearn}. The
hold-out is stratified by class where the class counts allow it, and ties go to
the smaller penalty. The same hold-out decisions score the shift
(Section~\ref{sec:contrib-imbalance}). Both routes keep the transform and the
proportions of the class weights fitted on all training series.

We fitted the iterative route beside the exact ridge on the 17
problems above the switch where the exact ridge still fits, each at its published
split. None of them has more than 9{,}456 training series. The two
routes give the same prediction for between 88\,\% and
100\,\% of a problem's test series. The paired mean difference in
accuracy, iterative minus exact, is $-0.4$ points, with a bootstrap
interval of $[-0.8, +0.1]$. The iterative route
wins on six problems and loses on eleven. Its largest loss is
2.7 points. A paired Wilcoxon test, without correction, finds the two
routes not significantly different in accuracy or in balanced accuracy. That
result does not show the routes equivalent. The claim is only that the iterative
route lies within about one point of the exact ridge on average.

\subsection{The streaming objective}
\label{app:stream-objective}

This subsection supports Equation~\ref{eq:stream} and Section~\ref{sec:contrib-scale}.

A step on series $s$ has size 1 for the coefficients and $10^{-2}$ for the
unpenalized intercept, scikit-learn's decay for sparse input. With the score
$f_c(X_s)$ taken before the step, the coefficients and the intercept follow
\begin{gather*}
\beta_c \leftarrow (1 - 10^{-5})\,\beta_c + \omega_{y_s}\big(\tau_{sc} -
f_c(X_s)\big)\,\phi(X_s), \\
\beta_{c,0} \leftarrow \beta_{c,0} + 10^{-2}\,\omega_{y_s}\big(\tau_{sc} - f_c(X_s)\big).
\end{gather*}
An epoch visits the chunks of the cache, whose rows stay fixed, in a new random
order, and shuffles the rows within each chunk. The penalty is per series, as in
scikit-learn. The fitted head is the mean of the coefficients and intercepts over
all $20n$ steps, as in scikit-learn's averaged mode. Bach and
Moulines~\cite{BachMoulines2013} bound the error of averaged constant-step
stochastic gradient descent for least squares when each series is used once and
the step is below 1 on rows of unit norm. This head takes a step of 1 and 20
passes, so their bound does not cover it. Summed over the series, the penalty is a
ridge penalty of $n \cdot 10^{-5}$. For 100~$\le n \le$~1{,}000{,}000 on balanced
classes, this penalty lies between 0.001 and 10, the ends of the ridge grid of
Table~\ref{tab:head}. Unbalanced classes raise this penalty in the batch head's
units, since $\omega$ is then $\gamma$ times a constant below 1.

The streaming head makes 20 passes over the cache for all classes together. A
step of the iterative route takes two passes over $\Phi$, shared by the classes of
a group, and the route solves twice. On the problems of
Appendix~\ref{app:ridge-routes}, at their published split and at eight further
splits of the four UCR problems, the iterative route makes 34 to
280 passes over $\Phi$ (72 at the median).

\end{document}